\documentclass{article}
\PassOptionsToPackage{numbers, sort&compress, comma, square}{natbib}
\usepackage[preprint]{neurips_2025}
\usepackage{natbib}

\usepackage[utf8]{inputenc} 
\usepackage[T1]{fontenc}    
\usepackage{hyperref}       
\usepackage{url}            
\usepackage{booktabs}       
\usepackage{amsfonts}       
\usepackage{nicefrac}       
\usepackage{microtype}      
\usepackage{xcolor}         
\usepackage{graphicx}
\usepackage{amssymb}
\usepackage{hyperref}
\usepackage{url}
\usepackage{xcolor}         
\usepackage{multirow}
\usepackage{graphicx}
\usepackage{color,colortbl}
\usepackage{pifont}
\usepackage{arydshln}
\usepackage{amsmath}
\usepackage{algorithm}
\usepackage{algpseudocode}
\usepackage{subcaption}
\usepackage{wrapfig}
\usepackage{enumitem}
\usepackage{stfloats}
\usepackage{amssymb}
\usepackage{booktabs}
\usepackage{tabularx}
\usepackage{makecell} 
\usepackage{booktabs}
\usepackage{adjustbox}
\usepackage{algorithm}
\usepackage{algorithmicx}
\usepackage{algpseudocode}
\usepackage{amsmath}  
\usepackage{wrapfig}
\usepackage{enumitem}

\title{Beyond One-Size-Fits-All Pruning via Evolutionary Metric Search for Large Language Models}

\author{Shuqi Liu$^{1,2}$, Bowei He$^{1}$, Han Wu$^{2,}$$^\dagger$, Linqi Song$^{1}$$^\dagger$\\
$^{1}$ Department of Computer Science, City University of Hong Kong\\
$^{2}$ Huawei Noah's Ark Lab\\
\texttt{\{shuqiliu4-c, boweihe2-c\}@my.cityu.edu.hk}\\
\texttt{wu.han1@huawei.com}\\
\texttt{linqi.song@cityu.edu.hk}
}

\begin{document}

\maketitle

\begin{abstract}
As large language models (LLMs) scale to hundreds of billions of parameters, post-training pruning has become crucial for efficient deployment. Current approaches, however, rely on static pruning metrics that fail to account for two fundamental variations: (1) weight distribution differences across model families, and (2) activation pattern variations across downstream tasks. This one-size-fits-all characteristic in current pruning metrics leads to systematically suboptimal performance when applied to diverse model-task combinations. 
To systematically investigate the relationships between optimal pruning metrics, model families, and downstream tasks, we propose \textsc{OptiShear}, a global evolutionary optimization framework that automatically discovers optimal pruning metrics for specific model-task pairs. Unlike conventional evaluation metric-based search paradigms, our method directly optimizes a global divergence objective. Thus, \textsc{OptiShear} not only improves computational efficiency but also ensures a global optimum compared to existing local pruning metrics.
Through comprehensive experiments with 16 LLMs (ranging from 125M to 70B parameters) and 10 diverse tasks, we characterize three fundamental properties of optimal pruning metrics for cross-model and cross-task scenarios: (1) cross-task generalization from complex to simpler tasks, (2) cross-model transferability from stronger to weaker models, and (3) improved alignment between transformed weights and activations—reducing their discrepancy by up to 90\% compared to prior methods.
\end{abstract}

{
\let\thefootnote\relax\footnotetext{
$^\dagger$Corresponding author.}
}
\section{Introduction} \label{sec:intro}

Large language models (LLMs) \cite{achiam2023gpt, touvron2023llama, le2023bloom} have demonstrated exceptional capabilities in language understanding and generation across various complex benchmarks \cite{bubeck2023sparks, wei2022emergent, wei2022chain}. However, their massive size poses significant challenges for inference and deployment due to extensive computational requirements.
Model pruning has emerged as a promising compression technique, which reduces model size by setting specific weights to zero. While traditional pruning approaches rely on retraining or iterative training to maintain performance \cite{lecun1989optimal, hassibi1993optimal, han2015learning, liu2018rethinking, blalock2020state, frankle2018lottery, renda2019comparing}, these methods become impractical for billion-parameter LLMs. Consequently, post-training pruning (PTP) has gained increasing attention due to its resource efficiency. PTP methods work by developing metrics to assess weight importance, allowing for the removal of less critical weights without the need for retraining \cite{frantar2023sparsegpt, sun2023simple, zhangplug}.
Nevertheless, most existing PTP metrics adopt a \textit{one-size-fits-all} strategy, where a single, fixed metric is used to determine weight importance across all models and tasks, without considering variations in weight distributions or input-dependent activation patterns.

\begin{figure*}[htbp]
  \centering  \includegraphics[width=1.0\textwidth]{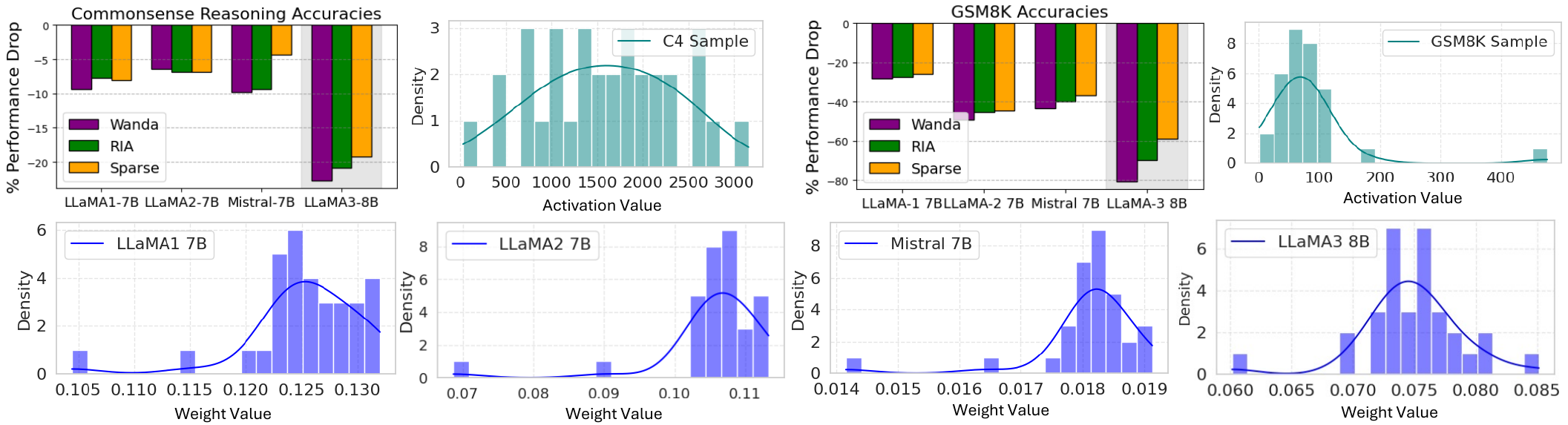}
  \caption{
  Existing pruning metrics, which follow a \textit{one-size-fits-all} design, fail on LLaMA-3 due to its distinct, highly concentrated weight distribution—marked by a symmetric bell-shaped curve with few near-zero weights—compared to the more dispersed patterns in LLaMA-1/2 and Mistral. 
  }
  \vspace{-1em}
  \label{fig:weight_distribution}
\end{figure*}

However, the assumption that measurement of \textit{important} weights remains consistent across different model architectures and tasks does not hold true in practice. As illustrated in Figure \ref{fig:weight_distribution}, employing fixed pruning metrics leads to systematic performance declines when applied to diverse model-task combinations.
For instance, applying SparseGPT, Wanda, and RIA \citep{frantar2023sparsegpt, sun2023simple, zhangplug} to the LLaMA-3 \citep{llama3} model results in significant accuracy degradation, with drops of 1.9× and 3.1× on the MMLU and GSM8K benchmarks, respectively, compared to its LLaMA-2 counterpart.
We trace this issue back to two fundamental components of existing pruning metrics: weight distributions and activation patterns \footnote{For better cross-layer visualization, we first compute sub-module-level weight $L_1$ norms and the sum of activation $L_2$ norms, then average across sub-modules to obtain a single value per layer. These values are used to plot histograms that capture overall distribution trends.}. The LLaMA-3 8B model exhibits a markedly different weight distribution compared to LLaMA-2, while the C4 (commonsense reasoning) and GSM8K (mathematical reasoning) datasets yield distinct activation patterns.
These findings highlight that fixed pruning metrics are inadequate for capturing the diversity present across models and tasks, ultimately resulting in suboptimal compression.


To systematically investigate how optimal pruning metrics depend on both model families and downstream tasks, we propose \textsc{OptiShear}, a global evolutionary optimization framework that automatically discovers optimal pruning metrics for specific model-task pairs. Our method introduces two key innovations: (1) a flexible metric space that combines weight magnitudes with input activations; and (2) a global divergence objective that evaluates pruning quality by measuring the final output discrepancy between the original and pruned models. 
Unlike conventional evaluation metric-based search paradigms, \textsc{OptiShear} not only improves computational efficiency but also ensures a global optimum compared to existing local pruning metrics.

We conduct extensive experiments on 16 large language models spanning diverse architectures—including OPT, Mistral, and the LLaMA-1/2/3 families—ranging in size from 125M to 70B parameters. Our evaluation covers 10 diverse downstream tasks, representing key domains such as language modeling, commonsense reasoning, and mathematical reasoning.
From these results, we uncover three key properties of optimal pruning metrics in cross-model and cross-task settings:
(1) Cross-task effectiveness: Pruning metrics discovered from complex reasoning tasks (e.g., GSM8K) generalize well to simpler tasks (e.g., language modeling).
(2) Cross-model transferability: Metrics discovered on stronger models (e.g., LLaMA-3) transfer effectively to weaker or related ones (e.g., Mistral).
(3) Improved weight-activation alignment: we find that effective pruning requires tight alignment between transformed weights and activations—\textsc{OptiShear} reduces their discrepancy by up to 90\% compared to existing methods.
These insights highlight the limitations of fixed, \textit{one-size-fits-all} pruning strategies and validate \textsc{OptiShear} as a principled framework for discovering adaptive, global, and task-aware compression policies.

\section{Related Work}
\label{rel_work}

\paragraph{Emergent Large Features of LLMs.}
Transformer-based large language models exhibit emergent features with exceptionally large magnitudes—both in weights and activations—occurring rarely but playing a critical role in information representation \citep{kovaleva2021bert, puccetti2022outliers, wei2022outlier, dettmers2022gpt3, sun2024massive}. These outlier features are essential for model performance, as ablating them significantly degrades accuracy \citep{dettmers2022gpt3, sun2024massive}. Their presence has influenced recent advances in quantization \citep{lin2023awq, dettmers2023spqr, xiao2023smoothquant} and pruning methods \citep{sun2023simple, zhangplug}, which explicitly account for such patterns. Building on this, our work further demonstrates that the interplay between weight magnitudes and activation outliers provides a powerful signal for identifying which weights to prune.

\paragraph{Post-Training Pruning.}
Post-training pruning (PTP) has become a popular approach for compressing large language models without retraining, offering efficiency gains in both computation and storage \citep{hubara2021accelerated, kwon2022fast, frantar2023sparsegpt}. Recent methods focus on defining weight importance metrics to identify and remove less critical parameters. Magnitude pruning \citep{han2015learning}, which removes weights with the smallest absolute values, is simple but often underperforms on LLMs. SparseGPT \citep{frantar2023sparsegpt} improves accuracy by solving a layer-wise reconstruction problem, though at increased computational cost. Wanda \citep{sun2023simple} simplifies this process by combining weight magnitude with input activation norms, while RIA \citep{zhangplug} further refines the metric with a relative importance coefficient. These one-shot methods have become strong baselines for efficient LLM pruning.

However, most existing techniques follow a local pruning paradigm—applying independent sparsity constraints per layer—which may neglect global model dependencies and lead to suboptimal performance under high sparsity. In contrast, recent work SparseLLM \citep{baisparsellm} introduces a global pruning framework that optimizes sparsity across the entire model, achieving better accuracy retention. Inspired by this shift, our method leverages the interaction between outlier weights and activations to guide global pruning decisions, enabling more effective and accurate model compression.

\section{\textsc{OptiShear}: Global Evolutionary Pruning Metric Discovery}

This section introduces \textsc{OptiShear}, a principled framework for automatically discovering pruning metrics that are tailored to specific model-task pairs. The method is structured in three stages:

\begin{itemize}
    \item Meta Pruning Metric: We define a flexible, adaptive metric space that combines weight magnitudes and input activations, allowing dynamic adjustment based on the statistical properties of each model.
    \item Global Divergence Objective: To evaluate the quality of candidate metrics, we define an objective function based on the global output divergence between the original and compressed models.
    \item Evolutionary Search Algorithm: Finally, we employ an evolutionary search algorithm to efficiently explore the metric space and identify optimal pruning strategies under sparsity constraints.
\end{itemize}


\subsection{Meta Pruning Metric: Adaptive Balancing of Weights and Activations}
\label{sec:metric}

While many existing pruning metrics combine weight magnitudes and input activations (e.g., Wanda \cite{sun2023simple} uses their product), they typically rely on fixed functional forms that do not adapt to the distinct characteristics of different model families or downstream tasks. This limits their effectiveness, especially given the diverse weight distributions and task-specific activation patterns observed in modern LLMs.
To address this limitation, we propose a meta pruning metric with two key components that enable flexible, data-driven adaptation:
(1) Normalization coefficients $\alpha(\cdot)$ and $\beta(\cdot)$, derived from global statistics of the weight matrix and activation matrix;
(2) Transformation functions $F_1(\cdot)$ and $F_2(\cdot)$, which allow for non-linear reweighting and scaling of raw values.

These additions enable our metric to dynamically balance the relative importance of weights and activations based on each model-task pair:
\begin{equation}
    S_{ij} = \alpha(|W_{ij}|) \cdot F_1(|W_{ij}|) \times \beta(\|X_j\|_2) \cdot F_2(\|X_j\|_2),
\label{eq:meta_metric}
\end{equation}
where $ |W_{ij}| $ denotes the absolute value of the weight connecting neuron $i$ to neuron $j$, $ \|X_j\|_2 $ is the $L_2$ norm of the input feature $j$ computed across tokens in the calibration dataset, $ \alpha(\cdot) $ and $ \beta(\cdot) $ are normalization functions that map raw values to scaling coefficients, and $ F_1(\cdot) $, $ F_2(\cdot) $ are transformation functions selected from a predefined set of nonlinear operations 

The full set of candidate operations for $\alpha(\cdot), \beta(\cdot)$ and $F_1(\cdot), F_2(\cdot)$ is summarized in Table~\ref{coe-ope}. These include various normalization strategies for coefficient computation and nonlinear transformations for value scaling. This diverse search space enables our meta metric to adaptively balance weight and activation importance across different model families and tasks. We refer readers to Table \ref{tab:coe_eqs} and Table\ref{tab:transformation_candidates} in Appendix \ref{app:method_details} for detailed definitions of the coefficient functions and candidate transformation operations used in our meta pruning metric. 

\begin{table*}[htbp]
\centering
\caption{Predefined coefficient and transformation function candidates for the Meta Pruning Metric.}
\label{coe-ope}
\vspace{-3pt}
\adjustbox{width=0.9\textwidth}{
\begin{tabular}{ll}
\toprule
\textbf{Coefficient Candidates ($\alpha, \beta$)} & Uniform Weighting, Frobenius Normalization \\
& Global Sum Scaling, Global Mean Scaling, Row-wise Normalization, \\
& Column-wise Normalization, Relative Magnitude Scaling \\
\midrule
\textbf{Transformation Candidates ($F_1, F_2$)} & Identity, Square, Square Root, Logarithm, Exponential, Sigmoid, Softmax \\
\bottomrule
\end{tabular}
}
\vspace{-8pt}
\end{table*}


\subsection{Objective Function: Global Divergence Minimization}
\label{sec:objective}

To guide the discovery of effective pruning strategies, we define a global divergence objective that measures the functional impact of pruning across the entire model. This contrasts with conventional pruning approaches, such as SparseGPT, Wanda, and RIA, which operate in a layer-wise manner and often neglect the cumulative effects of weight removal on overall model behavior.

Instead of minimizing local reconstruction errors independently for each layer, our objective directly captures how pruning affects the model's final output under task-specific inputs. As shown in recent work~\cite{baisparsellm}, such global objectives better preserve the alignment between compressed and original models.
Formally, we define our optimization objective as:
\begin{equation}
    \mathcal{L}_{\text{div}} = \mathbb{E}_{x \sim \mathcal{D}} \left[ \| \mathbf{h}_L(x) - \mathbf{h}_L^{\text{pruned}}(x) \|^2_2 \right],
    \label{eq:objective}
\end{equation}

where
$ \mathbf{h}_L(x) $ and $ \mathbf{h}_L^{\text{pruned}}(x) $ denote the final-layer hidden representations of the original and pruned models, respectively;
$ x \sim \mathcal{D} $ is sampled from a calibration dataset reflecting the input distribution of the target task.

This formulation ensures that pruning metrics reflect the model's overall behavior, rather than being driven by layer-wise approximations commonly used in existing methods. 
In the following section, we describe how this objective is integrated into an evolutionary search framework to automatically discover adaptive and task-aware pruning metrics.

\subsection{Search Strategy: Metric Space Exploration via Evolutionary Search}
\label{sec:search}

Given the flexible formulation of our meta pruning metric (Section~\ref{sec:metric}), we treat pruning metric discovery as a search problem over a space of adaptive configurations, where each candidate solution corresponds to a unique combination of normalization coefficients and transformation functions $(\alpha, \beta, F_1, F_2)$.
This space is discrete, high-dimensional, and non-differentiable—making gradient-based optimization infeasible and greedy heuristics ineffective. To navigate this complex landscape efficiently, we employ Non-dominated Sorting Genetic Algorithm II (NSGA-II) \cite{deb2002fast}, a well-established evolutionary algorithm known for its strong global search capabilities.

In our setup, each individual in the population encodes a full pruning strategy. We apply the corresponding metric across all layers, retaining the top-k\% weights per layer according to the target sparsity. The resulting pruned model is evaluated using our global divergence objective $\mathcal{L}_{\text{div}}$, defined in Section~\ref{sec:objective}, which serves as the fitness function for the evolutionary search.
Through iterative rounds of selection, crossover, and mutation, the algorithm evolves increasingly effective metric configurations (see Algorithm~\ref{alg:search} in Appendix \ref{app:method_details}for full details). This enables us to discover adaptive, task-aware pruning metrics without retraining or fine-tuning.

\section{Experiments}
\label{sec:experiments}

\subsection{Setup}

\paragraph{Models and Evaluations.}
We conduct extensive experiments on 16 large language models (LLMs) spanning multiple architectures—including OPT, Mistral, and the LLaMA-1/2/3 families—ranging in size from 125M to 70B parameters. Our evaluation covers 10 diverse downstream tasks, representing key domains such as language modeling, commonsense reasoning, and mathematical reasoning.
The model set includes variants of LLaMA-1 \citep{touvron2023llama} and LLaMA-2 \citep{touvron2023llama2} with sizes ranging from 7B to 70B, as well as LLaMA-3 8B \citep{llama3} and Mistral 7B \citep{jiang2023mistral}, including both base and instruction-tuned versions. We also evaluate on OPT models \citep{zhang2022optopenpretrainedtransformer} across scales from 125M to 13B.
For evaluation, we follow prior work \citep{sun2023simple, xia2023sheared} and assess performance on seven general knowledge tasks from the EleutherAI LM Harness \citep{eval-harness}, language modeling on the WikiText validation set \citep{merity2016pointer}, and the challenging benchmarks GSM8K \citep{cobbe2021training} (mathematical reasoning) and MMLU \citep{hendrycks2020measuring} (multi-task understanding).

\paragraph{Baselines \& Calibration data.}
We compare \textsc{OptiShear} with four established pruning baselines. These include: Magnitude Pruning \citep{han2015learning}, SparseGPT \citep{frantar2023sparsegpt}, Wanda\citep{sun2023simple}, and RIA \citep{zhangplug}. 
For calibration, we sample 128 sequences from the C4 training set for general knowledge and language modeling tasks (WikiText), and use 100 truncated examples (max 512 tokens) from each of GSM8K and MMLU.
We follow Wanda \citep{sun2023simple} and perform neuron-wise pruning, comparing weight importance scores within each output neuron. We evaluate three sparsity types: unstructured, semi-structured 4:8, and 2:4 sparsity \citep{sun2023simple,zhangplug}.

\begin{table*}[thbp]
\caption{Mean zero-shot accuracies (\%) on 7 general knowledge tasks (LM Harness), WikiText perplexity, GSM8K, and MMLU accuracies (\%) of pruned LLaMA-1/2/3 and Mistral models. Best results among pruning methods are \textbf{bolded}.}
\resizebox{\textwidth}{!}{
\fontsize{10}{12} \selectfont
\begin{tabular}{lccccccccccccc}
\toprule
\multirow{2}{*}{Method} &
  \multirow{2}{*}{\begin{tabular}[c]{@{}l@{}}Weight\\ Update\end{tabular}} &
  \multirow{2}{*}{Sparsity} &
  \multicolumn{2}{c}{LLaMA-1} & &
  \multicolumn{2}{c}{LLaMA-2} & &
  \multicolumn{2}{c}{LLaMA-3} & &
  \multicolumn{2}{c}{Mistral} \\
  \cline{4-5} \cline{7-8} \cline{10-11} \cline{13-14}
  &  &  & 7B & 13B & & 7B & 13B & & 8B & 8B-Inst & & 7B & 7B-Inst \\ \hline
  \rowcolor{gray!15}
\multicolumn{14}{c}{LM Harness Accuracy (\%)} \\
Dense & - & 0\%  & 59.70 & 62.58 & & 59.72 & 63.03 & & 64.21 & 64.15 & & 60.06 & 66.69  \\
\hdashline
Magnitude   & \ding{55} & 50\% & 46.89 &  47.34 & & 52.40 & 52.90 & & 44.87 & 45.31 &  & 57.24 &  63.34  \\
SparseGPT   & $\checkmark$ & 50\% &  54.86 & 58.54  && 55.90 & 60.70 & & 53.87  &  55.89 &  & 57.49 &  62.46  \\
Wanda       & \ding{55} & 50\% & 54.08 & 59.18 & & 55.89 & 60.88 && 49.66 & 51.34 & &  54.20  &  61.04  \\
RIA         & \ding{55} & 50\% &   \textbf{55.10}  &   59.45   & & 55.67 &   61.03 & &  50.76     &  50.64 &  &  54.39   &  60.48  \\
\hdashline
\textsc{OptiShear} & \ding{55} & 50\% &   \textbf{55.10}    &   \textbf{59.73}  &  &  \textbf{57.47} &  \textbf{61.42}  &  &  \textbf{55.50}     &  \textbf{55.94} & & \textbf{59.33} & \textbf{63.51} \\
\hline
\rowcolor{gray!15}
\multicolumn{14}{c}{WikiText Perplexity} \\
Dense & - & 0\%  & 5.37 & 4.80 & &  5.04 & 4.56 & & 5.80 & 7.91 & & 5.23 & 4.90   \\
\hdashline
Magnitude  & \ding{55} & 50\% & 13.27 & 13.55  & & 11.96 & 6.16 & & 73.93 & 5.5E2 & & 7.14 & 6.59   \\
SparseGPT  & $\checkmark$ & 50\% & 6.92 & 5.87 &  & 6.59 & 5.72 &&  10.89 & 13.27&  & 6.42 & 7.02   \\
Wanda  & \ding{55} & 50\% & 6.90 & 5.82 &  & 6.47 & 5.64 & & 10.57 & 16.37 & & 7.24 & 7.22   \\
RIA & \ding{55} & 50\% & 6.81 & 5.83  & & 6.43 & 5.63 & & 12.56 & 15.57 & & 7.27 & 7.21   \\
\hdashline
\textsc{OptiShear} & \ding{55} & 50\% & \textbf{6.78} & \textbf{5.74}   &&  \textbf{6.35} & \textbf{5.51} & & \textbf{9.23} & \textbf{11.37} && \textbf{6.22} &  \textbf{6.55}   \\
\hline
\rowcolor{gray!15}
\multicolumn{14}{c}{GSM8K Accuracy (\%)} \\
Dense & - & 0\%  & 11.07 & 17.82 & & 14.59 & 19.86 & & 52.39 & 74.45 & & 40.11  &  47.76  \\
\hdashline
Magnitude  & \ding{55} & 50\% & 1.52 & 5.99 && 2.05 & 6.22  & & 1.97 & 1.29 &  & 15.53  &  27.37  \\
SparseGPT  & $\checkmark$ & 50\% & 8.19 & 15.60 & & 8.11 & 13.42  & & 21.46 & 49.20 & & 25.40 &  33.97  \\
Wanda & \ding{55} & 50\% & 7.96 & 11.52  & & 7.43  & 9.10   & & 10.16 & 32.68 & & 22.74 &  33.59  \\
RIA & \ding{55} & 50\% & 8.04 & 11.14 & & 7.96 & 9.25  & & 15.85 & \textbf{52.39} & & 24.18 &  32.15  \\
\hline
\textsc{OptiShear} & \ding{55} & 50\% & \textbf{8.22} & \textbf{15.62}  & & \textbf{8.47} & \textbf{15.03} & & \textbf{43.07}  & 52.15 & & \textbf{25.78} &   \textbf{35.14}   \\
\hline
\rowcolor{gray!15}
\multicolumn{14}{c}{MMLU Accuracy (\%)} \\
Dense & - & 0\%  & 35.28 & 46.98 & & 41.97 & 51.47 & & 65.23 & 66.35 & & 58.92 & 62.54   \\
\hdashline
Magnitude  & \ding{55} & 50\% & 26.24 & 30.12 & & 26.04 & 43.83 & & 4.36 & 12.03 & & 50.83 & 49.52   \\
SparseGPT  & $\checkmark$ & 50\% & 29.48 & 38.29  & & 33.03 & 47.14 & & 49.50 & 52.27 & & 50.95 & 52.04   \\
Wanda & \ding{55} & 50\% & 29.81 & 37.84 & & 32.09 & 48.06 & & 49.05 & 53.15 & & 53.05 & 53.62   \\
RIA & \ding{55} & 50\% & 30.37 & 37.79 & & 31.46 & 47.39 & & 48.99 & 54.02 & & 52.67 & 53.14   \\
\hline
\textsc{OptiShear} & \ding{55} & 50\% & \textbf{31.05} & \textbf{39.76} & & \textbf{33.06} & \textbf{48.38} & & \textbf{51.22} & \textbf{55.60} &  & \textbf{53.87} &   \textbf{54.36}  \\
\bottomrule
\end{tabular}
}
\label{tab:all_res}
\end{table*}

\subsection{Main Results}
\paragraph{General Knowledge \& Language Modeling.}
As shown in Table~\ref{tab:all_res}, \textsc{OptiShear} consistently outperforms all baseline methods across LLaMA-1/2/3 and Mistral models on both general knowledge accuracy and WikiText perplexity. We provide detailed task-wise results for readers in Appendix \ref{appendix:lm-harness}.
On general knowledge tasks, our method achieves the highest scores across nearly all model variants. For example, on LLaMA-2 7B and 13B, \textsc{OptiShear} improves over RIA by +1.58\% and +0.39\%, respectively. More importantly, on LLaMA-3 and Mistral—where static pruning metrics (e.g., Wanda, RIA) suffer significant degradation—\textsc{OptiShear} surpasses even SparseGPT, a weight-update-based method, by up to +4.74\%. This highlights its ability to capture architecture-specific importance patterns without weight update.
In WikiText perplexity, \textsc{OptiShear} yields the lowest values among all pruned models. On LLaMA-3 8B, it reduces perplexity from 12.56 (RIA) to 9.23, a relative improvement of 26\%. In contrast, magnitude-based and static local pruning methods show large increases in perplexity after pruning, especially for LLaMA-3 and Mistral, indicating their unsuitability for these newer architectures.

\begin{table*}[thbp]
\caption{Performance of semi-structured N:M pruning on WikiText perplexity and GSM8K reasoning tasks for pruned LLaMA-1/2/3 and Mistral models.}
\resizebox{\textwidth}{!}{%
\fontsize{9}{11} \selectfont
\begin{tabular}{lccccccccccccccc}
\toprule
\multirow{2}{*}{Method} &
  \multirow{2}{*}{\begin{tabular}[c]{@{}l@{}}Weight\\ Update\end{tabular}} &
  \multirow{2}{*}{Sparsity} &
  \multicolumn{2}{c}{LLaMA-1} & &
  \multicolumn{2}{c}{LLaMA-2} & &
  \multicolumn{2}{c}{LLaMA-3} & &
  \multicolumn{2}{c}{Mistral} \\
\cline{4-5} \cline{7-8} \cline{10-11} \cline{13-14}
  &  &  & 7B & 13B & & 7B & 13B & & 8B & 8B-Inst & & 7B & 7B-Inst \\ \hline
\rowcolor{gray!15}
\multicolumn{14}{c}{WiKiText Perplexity} \\
Magnitude & \ding{55} & 4:8 & 17.48 & 16.80  & & 16.10 & 7.23 & & 2.5E2  &  5.6E2   &  & 8.78  &  8.67   \\
SparseGPT &  $\checkmark$  & 4:8 & 8.16  & 7.05   & & 7.89  &  6.54   &  & \textbf{15.57}  &  \textbf{16.62}   &  & 7.71  &   8.15  \\
Wanda & \ding{55} & 4:8 & 8.19  & 6.95 & & 8.01  & 6.60 & &  16.82  &  21.52   &  & 8.95  & 8.42   \\
RIA & \ding{55} & 4:8 & 8.18  & 6.97 & & 8.04  & 6.62 &  & 17.28  &  21.15   & &  8.91  &  8.51   \\
\hdashline
\textsc{OptiShear} & \ding{55} & 4:8 & \textbf{7.93}  & \textbf{6.65} & & \textbf{7.72}  & \textbf{6.34} & &  17.24  &  21.15    &  & \textbf{7.57}  & \textbf{7.66}   \\
\hline
Magnitude & \ding{55} & 2:4 & 49.06 & 19.33  & & 38.50 & 9.04 & &  5.3E3  &  5.3E3  &   &  13.18  &  11.83  \\
SparseGPT &  $\checkmark$  & 2:4 & 10.58 & 8.53 & & 10.38 & 8.26  && \textbf{23.43}  &  \textbf{26.68}  &  & 10.17  &  9.84    \\
Wanda & \ding{55} & 2:4 & 11.04 & 9.06 & & 11.31 & 8.46 & &  31.89  &   59.12  &  &  13.54   &  11.08  \\
RIA & \ding{55} & 2:4 & 11.10 & 9.24 & & 11.40 & 8.57 & &  31.79  &   38.00 &   &  13.61  &  11.21  \\
\hdashline
\textsc{OptiShear} & \ding{55} & 2:4 & \textbf{10.54} & \textbf{8.21}  &  & \textbf{10.34} & \textbf{7.97} & &  31.71  &  37.98  &  &  \textbf{10.13}  & \textbf{9.23}   \\
\hline
\rowcolor{gray!15}
\multicolumn{14}{c}{GSM8K} \\
Magnitude & \ding{55} & 4:8 & 1.53 & 3.48  & & 1.59 & 4.70 & &  4.16  &  7.81  &   &  9.60  &  14.15  \\
SparseGPT &  $\checkmark$  & 4:8 & 3.54 & 8.78 & & 4.84 & 8.20  && \textbf{9.23}  &  \textbf{18.35}  &  & 21.46  &  29.82    \\
Wanda & \ding{55} & 4:8 & 2.65 & 7.40 & & 3.10 & 8.13 & &  6.60  &   10.84  &  &  12.87   &  20.92  \\
RIA & \ding{55} & 4:8 & 3.17 & 8.74 & & 2.93 & 7.75 & &  8.12  &   17.59 &   &  17.36  &  27.18  \\
\hdashline
\textsc{OptiShear} & \ding{55} & 4:8 & \textbf{3.71}  & \textbf{9.29} & & \textbf{4.95}  & \textbf{8.53} & &  8.38  &  17.59    &  & \textbf{21.80}  & \textbf{30.39}   \\
\hline
Magnitude & \ding{55} & 2:4 & 0.74 & 2.29  & & 0.98 & 3.60 & &  0.24  &  3.12  &   &  3.80  &  9.26  \\
SparseGPT &  $\checkmark$  & 2:4 & 3.28 & \textbf{6.27} & & 3.10 & 6.53  && 1.71  &  \textbf{8.21}  &  & 7.52  &  19.45    \\
Wanda & \ding{55} & 2:4 & 2.75 & 6.12 & & 2.75 & 6.48 & &  2.27  &   3.51  &  &  4.93   &  10.79  \\
RIA & \ding{55} & 2:4 & 2.56 & 4.73 & & 2.79 & 5.65 & &  1.98  &   6.74 &   &  6.49  &  17.22  \\
\hdashline
\textsc{OptiShear} & \ding{55} & 2:4 & \textbf{3.34} & \textbf{6.27}  &  & \textbf{3.41} & \textbf{6.72} & &  \textbf{2.52}  &  6.74  &  &  \textbf{7.91}  & \textbf{20.33}   \\
\bottomrule
\end{tabular}%
}
\label{tab:N:M}
\vspace{-1em}
\end{table*}

\paragraph{Mathematic Reasoning \& Multi-Choice QA.}
We further evaluate the mathematic reasoning and multichoice QA capabilities of pruned models in GSM8K and MMLU. As shown in Table~\ref{tab:all_res}, \textsc{OptiShear} consistently outperforms all pruning baselines across model families, especially on LLaMA-3 and Mistral. 
On GSM8K, \textsc{OptiShear} achieves 41.17\% and 52.39\% accuracy on LLaMA-3 8B and 8B-Inst, respectively—surpassing Wanda and RIA by over 25\%. Notably, these results are obtained without any weight update during pruning, contrasting with SparseGPT’s reliance on such updates for competitive performance. This suggests that our framework better identifies task-relevant weights through its global divergence-based objective.
Similarly, for MMLU, \textsc{OptiShear} reaches up to 55.60\% accuracy on LLaMA-3 8B-Inst, setting a new state-of-the-art among post-training pruning methods. The relative gains increase on newer architectures, indicating that static importance metrics—designed for earlier model generations—are less effective in capturing critical structures in modern models. This aligns with observed differences in weight distributions (Figure~\ref{fig:weight_distribution}), further motivating the need for adaptive pruning strategies.

\vspace{-0.5em}
\paragraph{N:M Semi-Structured Pruning.}
While \textsc{OptiShear} is mainly designed for unstructured sparsity, it can be easily extended to semi-structured N:M sparsity \citep{mishra2021accelerating}, which can leverage NVIDIA’s sparse tensor cores to accelerate matrix multiplication in practice. 
Table\ref{tab:N:M} reports performance under 4:8 and 2:4 sparsity constraints on WikiText and GSM8K. We observe that \textsc{OptiShear} generally performs on par with or better than existing baseline methods across most model families, with the exception of LLaMA-3.
One possible explanation is that LLaMA-3, having been trained on a larger and more diverse dataset, exhibits higher knowledge density \citep{llama3}. As a result, semi-structured pruning—which removes continuous parameter blocks—may disproportionately impact critical representations, leading to greater performance degradation compared to unstructured approaches. This suggests that weight updates, as used in methods like SparseGPT, may be more crucial for maintaining performance on high knowledge density models.

\begin{table*}[thbp]
\caption{WikiText perplexity and per-task zero-shot accuracies (\%) on 7 general knowledge tasks for 50\% unstructured pruning on LLaMA-1 30B and LLaMA-2 70B models.}
\vspace{-0.5em}
\resizebox{\textwidth}{!}{%
\fontsize{9}{11} \selectfont
\begin{tabular}{ll|c|cccccccc}
\toprule
Model & Method & WikiText & BoolQ & RTE & HellaSwag & WinoGrande & ARC-e & ARC-c & OBQA & Average \\
  \hline
\multirow{6}{*}{\begin{tabular}[c]{@{}l@{}}LLaMA\\ 30B\end{tabular}} &
  Dense & 4.77 & 82.69 & 66.79 & 63.35 & 75.69 & 80.30 & 52.82 & 36.00 & 65.38 \\
  \cdashline{2-11}
 &
  Magnitude & 7.55 & 64.34 & 50.18 & 50.59 & 66.54 & 72.39 & 43.77 & 29.00 & 53.83 \\
 &
  SparseGPT &5.32&82.32&62.45&59.15&75.22&78.96&48.56&\textbf{35.00}&63.09
   \\
 &
  Wanda &5.98&81.90&65.34&\textbf{60.93}&\textbf{73.48}&\textbf{79.29}&49.66&34.60&63.60
   \\
 &
  RIA &5.16&\textbf{83.36}&67.15&60.01&72.85&78.70&48.29&33.60&63.42
   \\
 &
  \textsc{OptiShear} &\textbf{5.10}&\textbf{83.36}&\textbf{67.51}&\textbf{60.93}&72.61&78.91&\textbf{49.74}&34.20&\textbf{63.89}
   \\
   \hline
\multirow{6}{*}{\begin{tabular}[c]{@{}l@{}}LLaMA2\\ 70B\end{tabular}} &
  Dense &3.12&83.40 & 67.87& 66.10& 78.06& 82.55& 54.44& 37.20&67.08
  \\
  \cdashline{2-11}
 &
  Magnitude &4.98&70.55 & 60.65& 61.50& 73.48& 75.70& 49.23& 35.40&60.93 \\
 &
  SparseGPT &3.98&\textbf{83.55} & 70.40& 63.80& \textbf{78.85}& \textbf{82.40}& \textbf{53.75}& 38.20&67.28 \\
 &
  Wanda &3.99&82.50 & \textbf{73.65}& \textbf{64.10}& 78.14& 80.80& 52.65& 37.40&67.03 \\
 &
  RIA &3.91&83.25 & 71.49& 64.05& 77.74& 81.20& 53.16& 36.60&66.77 \\
 &
  \textsc{OptiShear} &\textbf{3.86}&83.25 & 73.21& 64.00& 78.48& 81.25& 53.07& \textbf{38.40}&\textbf{67.38} \\
  \bottomrule
\end{tabular}%
}
\label{tab:70b}
\end{table*}

\vspace{-0.5em}
\paragraph{Effectiveness on Larger Language Models.}
\label{sec:70b}
We further evaluate the scalability of \textsc{OptiShear} on two large models: LLaMA-1 30B and LLaMA-2 70B. As shown in Table~\ref{tab:70b}, \textsc{OptiShear} generally achieves competitive or better performance than existing methods across both WikiText perplexity and individual zero-shot tasks, without requiring weight updates. For LLaMA-1 30B, \textsc{OptiShear} obtains the lowest perplexity (5.10) among all pruning baselines and performs on par with or better than others on most zero-shot tasks. On LLaMA-2 70B, it also shows strong performance, achieving a perplexity of 3.86—close to that of the dense model (3.12)—and maintaining a higher average accuracy.
It is observed that the relative gains of \textsc{OptiShear} are somewhat smaller on these large-scale models compared to their smaller counterparts (e.g., 7B or 13B). This may be attributed to smaller models, with fewer parameters and simpler architectures, are more sensitive to pruning as the removal of key connections can severely disrupt information flow. Accordingly, our pruning metric achieves greater performance gains on smaller models by preserving critical weights.



\vspace{-0.5em}
\paragraph{Effectiveness on Other Model Architectures.}
We further evaluate the generalization of \textsc{OptiShear} across a wide range of OPT models, spanning from 125M to 13B parameters. Results on WikiText perplexity and average zero-shot accuracy over 7 general knowledge tasks (Table\ref{tab:opt_part_1}) shows that \textsc{OptiShear} consistently outperforms Wanda, RIA, and SparseGPT across nearly all model scales. In particular, it achieves the highest average task accuracy in most settings while maintaining competitive or lower perplexity—demonstrating both effectiveness and robustness for post-training pruning. Additional results for other model sizes are provided in Appendix \ref{appendix:lm-harness} (Table \ref{tab:opt_part_2}).

\begin{table*}[htbp]
\centering
\caption{WikiText perplexity and mean zero-shot accuracies (\%) on 7 general knowledge tasks of pruned OPT models (125M to 13B) with 50\% sparsity.}
\label{tab:opt_part_1}
\resizebox{\textwidth}{!}{%
\begin{tabular}{lcccccccccc}
\toprule
\textbf{Model} & \textbf{Method} & \textbf{PPL} & \textbf{BoolQ} & \textbf{RTE} & \textbf{HellaSwag} & \textbf{WinoGrande} & \textbf{ARC-e} & \textbf{ARC-c} & \textbf{OBQA} & \textbf{Avg. Acc.} \\
\midrule

\multirow{4}{*}{OPT-125M} 
& SparseGPT   & 37.11 & 56.39 & \textbf{52.76} & \textbf{29.42} & 52.49 & \textbf{37.84} & \textbf{17.49} & \textbf{12.40} & 36.97 \\
& Wanda       & 38.86 & 42.45 & 52.71 & 27.98 & 51.62 & 35.40 & 16.64 & 12.00 & 34.11 \\
& RIA         & 38.97 & 40.61 & 52.35 & 27.63 & 52.25 & 34.85 & 17.06 & 12.20 & 33.85 \\
& \textsc{OptiShear}       & \textbf{36.81} & \textbf{58.78} & 52.71 & 29.11 & \textbf{52.96} & 36.62 & 17.41 & \textbf{12.40} & \textbf{37.14} \\
\midrule

\multirow{4}{*}{OPT-2.7B} 
& SparseGPT   & 13.51 & 62.23 & 52.35 & \textbf{40.69} & \textbf{56.75} & 54.59 & 24.49 & 18.40 & 44.27 \\
& Wanda       & 14.39 & 62.26 & \textbf{52.71} & 32.08 & 50.99 & 44.19 & 18.69 & 14.40 & 39.33 \\
& RIA         & 14.02 & 62.29 & \textbf{52.71} & 31.78 & 50.83 & 44.11 & 19.11 & 14.80 & 39.38 \\
& \textsc{OptiShear}       & \textbf{13.17} & \textbf{63.79} & 51.26 & 40.45 & 56.27 & \textbf{55.30} & \textbf{24.83} & \textbf{18.80} & \textbf{44.39} \\
\midrule

\multirow{4}{*}{OPT-13B} 
& SparseGPT   & 11.28 & 61.74 & 57.40 & 45.06 & 63.06 & \textbf{62.54} & 29.18 & 21.80 & 48.68 \\
& Wanda       & 11.56 & \textbf{65.63} & 53.07 & 37.17 & 56.99 & 52.10 & 22.61 & 16.40 & 43.42 \\
& RIA         & 11.43 & 64.95 & 52.71 & 36.80 & 57.85 & 53.03 & 22.53 & 16.40 & 43.47 \\
& \textsc{OptiShear}       & \textbf{11.15} & 62.45 & \textbf{59.57} & \textbf{48.17} & \textbf{64.33} & 61.83 & \textbf{29.61} & \textbf{22.80} & \textbf{49.82} \\
\bottomrule
\end{tabular}
}
\end{table*}

\section{In-depth Analysis}
\label{sec:analysis}

\subsection{Behavior of Pruning Metrics Across Models and Tasks}

Our experimental findings reveal three fundamental properties of the pruning metrics discovered by \textsc{OptiShear}, shedding light on the characteristics of effective post-training pruning across diverse models and tasks.

\begin{figure*}[htbp]
  \centering  \includegraphics[width=0.9\textwidth]{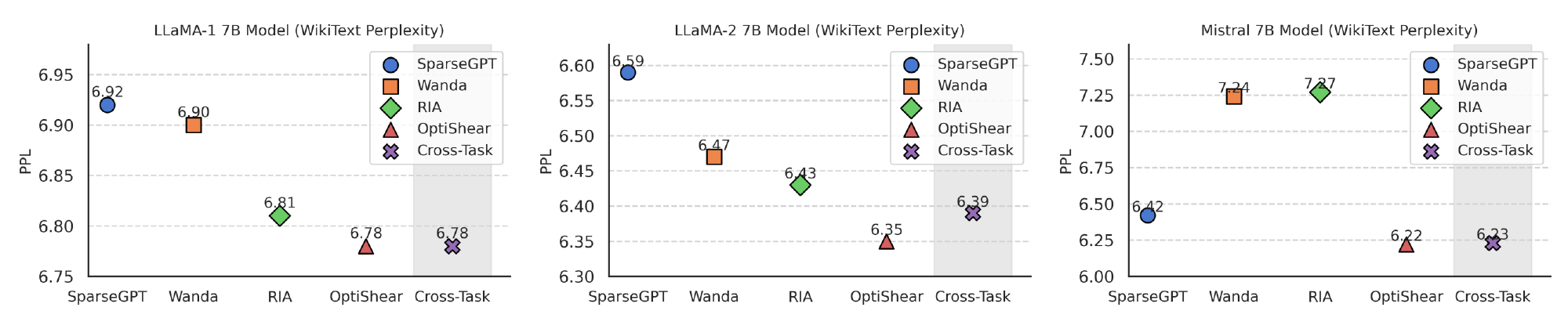}
  \caption{WikiText perplexity of metrics transferred from GSM8K (mathematical reasoning) to language modeling, showing strong cross-task generalization.}
  \vspace{-1em}
  \label{fig:cross_task_transfer}
\end{figure*}

\paragraph{Cross-task Generalization.}
We evaluate whether pruning metrics discovered on one task can generalize to others. Encouragingly, metrics learned from complex reasoning tasks (e.g., GSM8K) transfer well to simpler ones such as language modeling. As shown in Figure~\ref{fig:cross_task_transfer}, on LLaMA-1 7B WikiText perplexity, the GSM8K-derived metric achieves 6.78, outperforming SparseGPT (6.92), Wanda (6.90), and RIA (6.90). Similar gains are observed on LLaMA-2 and Mistral models (6.35 and 6.22, respectively).
In contrast, transferring metrics from simpler tasks (e.g., WikiText) to complex ones consistently leads to performance drops. 

\vspace{-1em}
\paragraph{Cross-model Transferability.}
\begin{wrapfigure}{r}{0.6\textwidth}
  \centering
  \includegraphics[width=\linewidth]{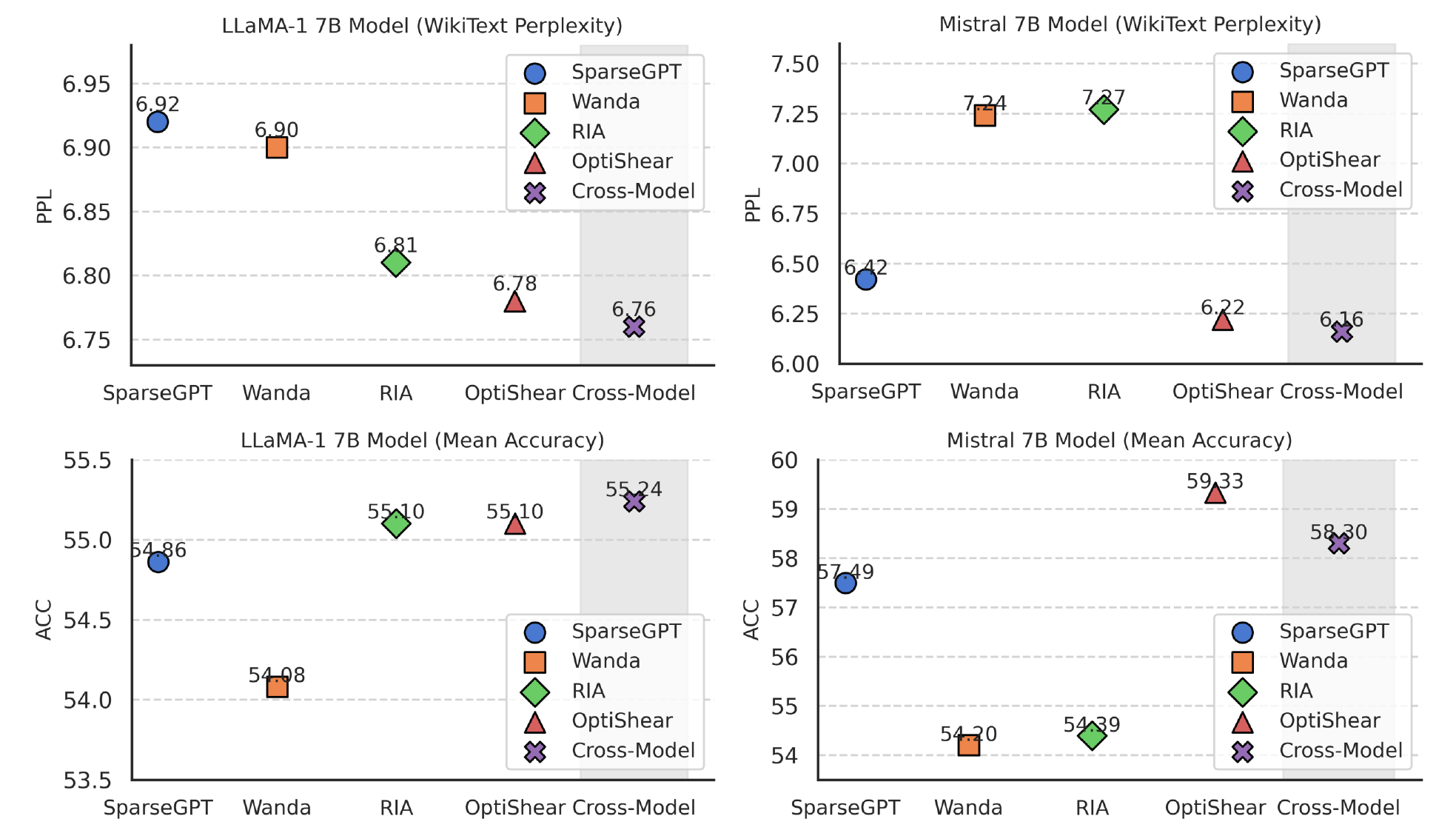}
  \caption{WikiText perplexity of metrics transferred from LLaMA-2 and LLaMA-3 to LLaMA-1 and Mistral models, showing strong cross-model transferability.}
  \label{fig:cross_model_transfer}
  \vspace{-1em}
\end{wrapfigure}

We further evaluate whether pruning metrics discovered on one model family can generalize to others. Specifically, we derive metrics from LLaMA-2 7B and LLaMA-3 8B—models known for strong reasoning performance—and apply them to LLaMA-1 and Mistral models. As shown in Figure~\ref{fig:cross_model_transfer},
applying LLaMA-3 metric to Mistral 7B outperforms both static baselines and even the in-task \textsc{OptiShear} configuration. Similarly, transferring the LLaMA-2 metric to LLaMA-1 achieves nearly the same performance as the task-specific metric. These results indicate that metrics discovered from stronger models can transfer effectively to weaker and related ones. 

\vspace{-1em}
\paragraph{Improved Weight-Activation Alignment.}
\begin{wraptable}{r}{0.55\textwidth}
    \centering
    \vspace{-1em}
    \caption{Layer-wise alignment between transformed weights and activations on C4 data. Lower indicates better alignment.}
    \resizebox{\linewidth}{!}{%
    \begin{tabular}{lccc}
        \toprule
        \textbf{Method} & \textbf{LLaMA-2 7B} & \textbf{LLaMA-3 8B} & \textbf{Mistral 7B} \\
        \midrule
        Wanda            & 82.66              & 79.31              & 392.13 \\
        RIA              & 22.34              & 21.15              & 44.77 \\
        \textsc{OptiShear} & \textbf{1.09}      & \textbf{0.1263}    & \textbf{0.0304} \\
        \midrule
        \% Reduction vs. Wanda & 98.7\% & 99.8\% & 99.99\% \\
        \% Reduction vs. RIA   & 95.1\% & 99.4\% & 99.93\% \\
        \bottomrule
    \end{tabular}
    }
    \label{tab:average_difference}
    \vspace{-1em}
\end{wraptable}

One key insight from our analysis is that effective pruning critically depends on the alignment between weight and activation importance. To examine this, we decompose several representative metrics—Wanda, RIA, and the one discovered by \textsc{OptiShear}—into their transformed weight and activation components.
For each layer, we measure the discrepancy by summing the transformed values within each linear sub-module, and then computing the average absolute difference across all sub-module pairs in that layer.

As shown in Table~\ref{tab:average_difference}, \textsc{OptiShear} significantly reduces this discrepancy—bringing it close to zero—compared to Wanda and RIA. 
This suggests that the adaptive scaling operations in \textsc{OptiShear} align weights with their activations, leading to more balanced and effective pruning. 
Combined with the results in Table~\ref{tab:all_res}, this supports our conclusion: better alignment between weights and activations yields more robust and effective model compression. We provide fine-grained layer-wise analysis of weight-activation alignment in Appendix~\ref{sec:w_x}.


\vspace{-1em}
\paragraph{Most Important Pruning Metrics.}
Given the observed cross-task generalization and cross-model transferability of pruning metrics, we identify that for LLaMA-1 and LLaMA-2 models, the most effective pruning strategy is derived from the metric learned on LLaMA-2 using GSM8K calibration data. This metric takes the following form:
\begin{equation}
    S_{ij} = ||W_{ij}||_{F}^{-1} \cdot |W_{ij}| \times ( 1/ \sum_{j} \|X_j\|_2)  \cdot (\|X_j\|_2)^{1/2},
\label{eq:important_1}
\end{equation}
In contrast, for LLaMA-3 and Mistral models, the best-performing metric is obtained from the search conducted on LLaMA-3 using the same calibration set, and is expressed as:
\begin{equation}
    S_{ij} =  (mn / \sum_{i,j} |W_{ij}|) \cdot |W_{ij}| \times (1 / \sum_{j} \|X_j\|_2 ) \cdot (\|X_j\|_2)^{1/2},
\label{eq:important_2}
\end{equation}

\vspace{-1em}
\paragraph{Comparison with Zero-Pruner.}
\begin{wrapfigure}{r}{0.4\textwidth}
  \centering
  \includegraphics[width=\linewidth]{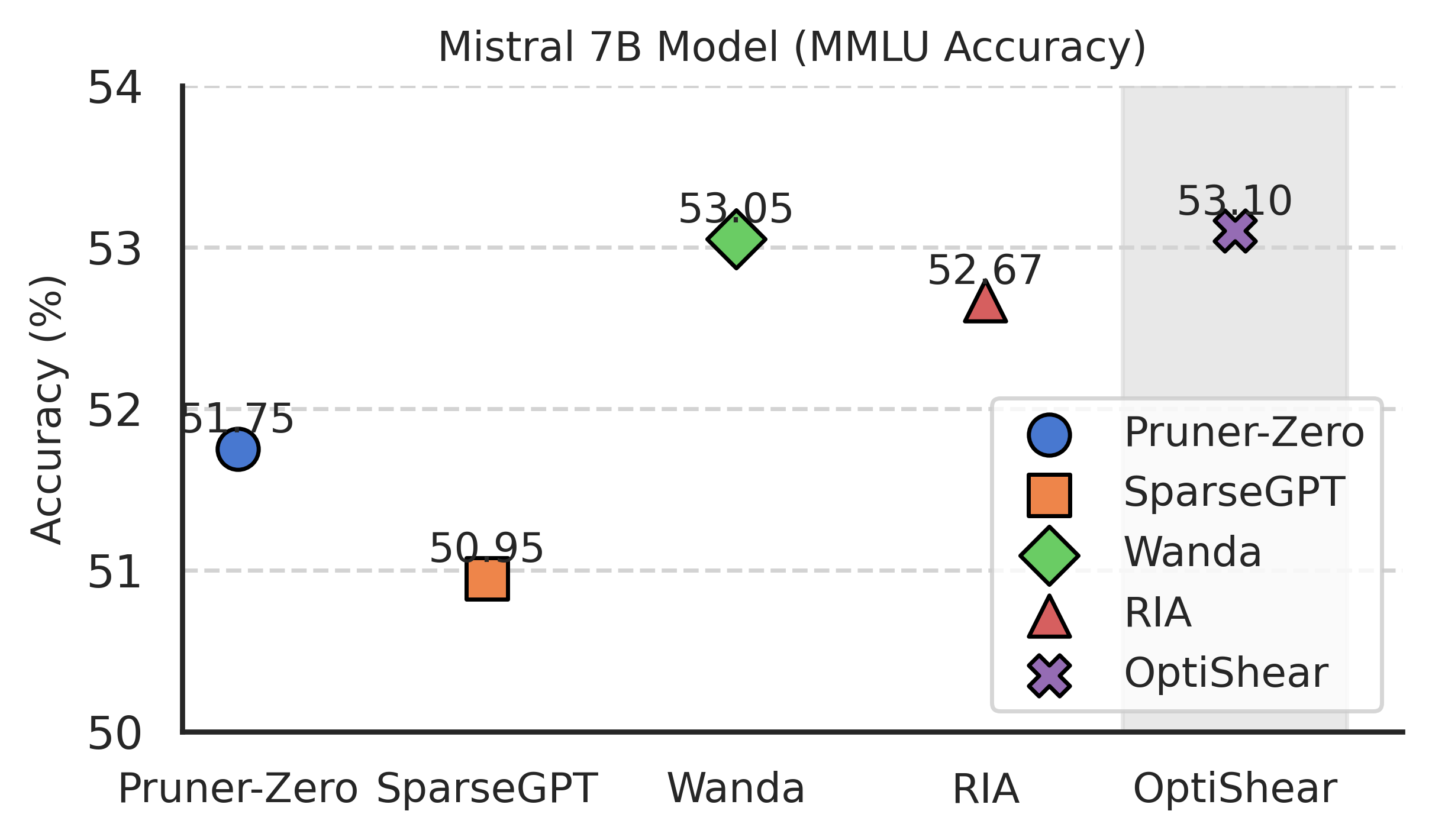}
  \caption{Comparison of pruning methods on Mistral 7B.}
  \label{fig:pruner_zero}
  \vspace{-1em}
\end{wrapfigure}

We further compare our method with Zero-Pruner \cite{dongpruner}, which also uses a genetic algorithm to discover pruning metrics. However, Zero Pruner relies on weight-gradient interactions and evaluates metrics using validation perplexity—both of which are computationally costly.
In contrast, our method leverages the interplay between weights and activations and uses a global divergence objective that requires only a single forward pass. As shown in Figure~\ref{fig:pruner_zero}, our metric outperforms gradient-based alternatives across the MMLU task on Mistral 7B, confirming the advantage of activation-aware pruning strategies.

\subsection{Evolutionary Search Process Analysis}

\paragraph{Search Time \& Cost.}
Table\ref{tab:time} summarizes the computational cost of searching for optimal pruning metrics across LLaMA-1/2/3 and Mistral models on a single NVIDIA RTX A6000 GPU. Each search run consists of 350 evolutionary trials, a number empirically found to be sufficient for convergence, as shown in Appendix \ref{app:method_details} Figure \ref{fig:ntrials}. As indicated in the table, the total search time remains under 2.5 GPU hours for all model variants, with most models completing within 1.5 hours. When distributed across multiple GPUs, the search typically finishes in less than an hour, making the overall computational footprint moderate and suitable for practical deployment scenarios.

\vspace{-1em}
\begin{table}[htbp]
\caption{Time cost of searching for optimal pruning metrics on LLaMA-1/2/3 and Mistral models.}
    \centering
    \setlength{\tabcolsep}{4.5pt}
    \resizebox{\textwidth}{!}{%
    \begin{tabular}{lcccccccc}
    \toprule
         Model & L1-7B & L1-13B & L2-7B & L2-13B & L3-8B & L3-8B-it & M-7B & M-7B-it \\
         \hline
         Metric Search (h:m:s) & 1:10:28 & 2:13:06 & 1:06:14 & 2:11:55 & 1:30:47 & 1:31:51 & 1:14:22 & 1:15:54 \\
         \bottomrule
    \end{tabular}
    }
    \label{tab:time}
\end{table}

\vspace{-1em}
\paragraph{Search Algorithms \& Robustness Analysis.}
\label{sec:robustness}
We conduct a robustness analysis using five search algorithms, including random search \citep{bergstra2012random}, the Tree-structured Parzen Estimator (TPE) \citep{bergstra2011algorithms, bergstra2013making, ozaki2022multiobjective}, and Quasi-Monte Carlo (QMC) \citep{bergstra2012random} sampler. Table~\ref{tab:robust} reports mean and standard deviation results across four benchmarks for pruned LLaMA-2 7B models. Each configuration is evaluated under three independent runs with different random seeds to ensure reliable comparisons. We report the performance outcomes of the NSGA-II search method in the main paper, as it generally outperforms other algorithms. 

\begin{table*}[htbp]
\centering
\caption{Statistical results of different search algorithms on LLaMA-2 7B model. 
We report the mean and standard deviation under 3 search process runs.}
\tiny
\resizebox{0.9\textwidth}{!}{%
\begin{tabular}{lcccc}
\toprule
Dataset        & Random & TPE & QMC  & NSGA-II \\
\hline
WikiText       & 6.89 ($\pm$ 0.0671)   & \textbf{6.33} ($\pm$ 0.0714) & 6.39 ($\pm$ 0.0700) & 6.35 ($\pm$ 0.0640) \\
GSM8K          & 7.96 ($\pm$ 0.2406)   & 8.33 ($\pm$ 0.2498)         & 8.08 ($\pm$ 0.2220) & \textbf{8.49} ($\pm$ 0.2646) \\
MMLU           & 31.11 ($\pm$ 0.3962)  & 31.06 ($\pm$ 0.4017)        & 31.80 ($\pm$ 0.4400) & \textbf{33.06} ($\pm$ 0.4687) \\
LM-harness     & 55.32 ($\pm$ 0.5300)  & 55.74 ($\pm$ 0.5367)        & 56.19 ($\pm$ 0.5234) & \textbf{57.47} ($\pm$ 0.5718) \\
\bottomrule
\end{tabular}%
}
\label{tab:robust}
\end{table*}

\section{Conclusion}
In this work, we presented \textsc{OptiShear}, a global evolutionary framework for discovering adaptive pruning metrics tailored to specific model-task pairs. By directly optimizing a global divergence objective, our method overcomes the limitations of static, one-size-fits-all pruning metrics that fail to account for variations in weight distributions across models and activation patterns across tasks. Extensive experiments across 16 LLMs and 10 diverse tasks demonstrate that \textsc{OptiShear} not only achieves superior compression performance but also reveals fundamental properties of effective pruning—such as cross-task generalization, cross-model transferability, and improved alignment between weights and activations.

\small
\bibliography{neurips_2025}







\appendix

\section{Technical Appendices and Supplementary Material}
\label{sec:appendices}

In this section, we provide supplementary details on experimental results and methodological choices, aimed at offering deeper insights into the empirical behavior and implementation of \textsc{OptiShear}.

\subsection{Additional Experimental Results}
\label{app:additional_results}

We present extended results across multiple models and tasks to further validate the generalization and robustness of our pruning framework.

\paragraph{Task-wise Results on LM Harness}
\label{appendix:lm-harness}
For LM-harness results, the 7 evaluated zero-shot tasks are: BoolQ \citep{clark2019boolq},
RTE \citep{wang2018glue}, HellaSwag \citep{zellers2019hellaswag}, WinoGrande \citep{sakaguchi2021winogrande}, ARC Easy and Challenge \citep{clark2018think}, and OpenbookQA \citep{mihaylov2018can}. For
reproducibility, we used v0.4.0 release. All tasks were evaluated on task version 0 except for BoolQ on task version 1. We show the task-wise performance of mean zero-shot accuracies of pruned LLaMA-1/2/3, Mistral models and OPT models in Tables \ref{tab:zero-1-7}, \ref{tab:zero-1-13}, \ref{tab:zero-2-7}, \ref{tab:zero-2-13}, \ref{tab:zero-3-8}, \ref{tab:zero-3-8-it}, \ref{tab:zero-m-1}, \ref{tab:zero-m-2}, \ref{tab:opt_part_2}. 

\begin{table*}[htbp]
    \centering
    \caption{Accuracies (\%) of LLaMA-1 7B model for 7 zero-shot tasks with unstructured 50\% sparsity.}
    \resizebox{\textwidth}{!}{%
    \begin{tabular}{lcccccccc}
         \toprule
         Method & BoolQ & RTE & HellaSwag & WinoGrande & ARC-e & ARC-c & OBQA & Average\\
         \hline
         Dense & 75.06 & 66.23 & 56.93 & 69.54 & 74.82 & 41.02 & 34.30 & 59.70\\
         Magnitude & 55.10 & 54.51 & 45.49 & 59.10 & 58.65 & 32.97 & 22.40 & 46.89 \\
         SparseGPT & 72.03 & 54.15 & 51.43 & 67.87 & 71.39 & 37.54 & 29.60 & 54.86 \\
         Wanda & 71.04 & 54.51 & 51.93 & 65.90 & 69.40 & \textbf{36.95} & 28.80 & 54.08\\
         RIA & 72.84 & 57.76 & 51.93 & 66.85 & \textbf{70.50} & 36.43 & 29.40 & 55.10\\
         Pruner-Zero & 70.28 & 56.68 & 47.27 & 64.96 & 66.92 & 33.25 & 26.80 & 52.31\\
         \hdashline
         Our Metric & \textbf{72.87} & 57.40 & 51.91 & \textbf{67.25} & 70.33 & 36.35 & 29.60 & 55.10\\
         GSM8K Metric & 71.04 & \textbf{58.48} & 52.39 & 67.17 & 69.91 & 37.46 & \textbf{30.20} & \textbf{55.24}\\
         LLaMA2 Metric & 70.73 & 57.63 & \textbf{53.24} & 67.01 & 70.24 & 37.97 & \textbf{30.20} & 55.15\\
         \bottomrule
    \end{tabular}
    }
    \label{tab:zero-1-7}
\end{table*}

\begin{table*}[htbp]
    \centering
    \caption{Accuracies (\%) of LLaMA-1 13B model for 7 zero-shot tasks with unstructured 50\% sparsity.}
    \resizebox{\textwidth}{!}{%
    \begin{tabular}{lcccccccc}
         \toprule
         Method & BoolQ & RTE & HellaSwag & WinoGrande & ARC-e & ARC-c & OBQA & Average\\
         \hline
         Dense & 78.03 & 70.51 & 59.63 & 72.89 & 77.28 & 46.55 & 33.20 & 62.58\\
         Magnitude & 55.19 & 52.23 & 43.65 & 63.36 & 57.82 & 32.53 & 26.60 & 47.34 \\
         SparseGPT & \textbf{76.89} & 60.95 & 54.99 & 71.46 & 72.15 & 42.17 & 31.20 & 58.54 \\
         Wanda & 75.73 & 62.48 & 55.70 & 71.68 & 72.91 & 43.45 & 32.20 & 59.18\\
         RIA & 76.44 & 62.34 & 56.13 & 72.73 & 72.42 & \textbf{43.87} & 32.20 & 59.45\\
         Pruner-Zero & 73.91 & 62.36 & 52.65 & 69.41 & 70.83 & 41.62 & 28.80 & 57.08\\
         \hdashline
         Our Metric & 76.67 & 62.45 & 56.11 & \textbf{73.63} & 73.25 & 43.62 & \textbf{32.40} & \textbf{59.73}\\
         GSM8K Metric & 76.62 & \textbf{62.89} & 55.48 & 72.79 & 72.58 & 43.78 & 32.00 & 59.45\\
         LLaMA2 Metric & 76.51 & 62.32 & \textbf{56.43} & 71.82 & \textbf{73.39} & 43.84 & \textbf{32.40} & 59.53\\
         \bottomrule
    \end{tabular}
    }
    \label{tab:zero-1-13}
\end{table*}

\begin{table*}[htbp]
    \centering
    \caption{Accuracies (\%) of LLaMA-2 7B model for 7 zero-shot tasks with unstructured 50\% sparsity.}
    \resizebox{\textwidth}{!}{%
    \begin{tabular}{lcccccccc}
         \toprule
         Method & BoolQ & RTE & HellaSwag & WinoGrande & ARC-e & ARC-c & OBQA & Average\\
         \hline
         Dense & 77.74 & 62.82 & 57.14 & 69.14 & 76.35 & 43.43 & 31.40 & 59.72\\
         Magnitude & 62.57 & 52.35 & 52.99 & 65.35 & 67.97 & 37.20 & 28.40 & 52.40 \\
         SparseGPT & \textbf{75.78} & 57.75 & 52.90 & \textbf{69.14} & 71.34 & 37.97 & 26.60 & 55.90 \\
         Wanda & 75.35 & 53.43 & 52.63 & 67.25 & \textbf{72.35} & 39.42 & 30.80 & 55.89\\
         RIA & 75.66 & 53.79 & 52.25 & 67.25 & 72.05 & 37.71 & \textbf{31.00} & 55.67\\
         Pruner-Zero & 73.48 & 53.29 & 49.18 & 65.83 & 69.92 & 38.36 & 26.60 & 53.81\\
         \hdashline
         Our Metric & 74.62 & \textbf{62.82} & \textbf{57.14} & 68.03 & 71.00 & 38.91 & 29.80 & \textbf{57.47}\\
         GSM8K Metric & 75.11 & 53.79 & 53.55 & 67.25 & 72.31 & \textbf{39.93} & 30.40 & 56.05\\
         LLaMA2 Metric & 75.11 & 53.79 & 53.55 & 67.25 & 72.31 & \textbf{39.93} & 30.40 & 56.05\\
         \bottomrule
    \end{tabular}
    }
    \label{tab:zero-2-7}
\end{table*}

\begin{table*}[htbp]
    \centering
    \caption{Accuracies (\%) of LLaMA-2 13B model for 7 zero-shot tasks with unstructured 50\% sparsity.}
    \resizebox{\textwidth}{!}{%
    \begin{tabular}{lcccccccc}
         \toprule
         Method & BoolQ & RTE & HellaSwag & WinoGrande & ARC-e & ARC-c & OBQA & Average\\
         \hline
         Dense & 80.52 & 65.34 & 60.33 & 71.95 & 79.38 & 48.47 & 35.20 & 63.03\\
         Magnitude & 57.62 & 55.87 & 54.53 & 65.85 & 70.47 & 38.13 & 27.80 & 52.90 \\
         SparseGPT & 81.42 & 65.26 & 55.83 & 72.64 & 74.91 & 42.23 & \textbf{32.60} & 60.70 \\
         Wanda & 81.86 & 64.08 & 56.92 & 71.37 & 76.12 & 43.81 & 32.00 & 60.88\\
         RIA & \textbf{81.93} & 64.02 & 57.73 & 71.89 & 76.24 & 43.46 & 32.00 & 61.03\\
         Pruner-Zero & 77.86 & 61.22 & 56.89 & 67.90 & 74.16 & 39.81 & 29.40 & 58.18\\
         \hdashline
         Our Metric & 80.97 & \textbf{66.17} & 59.68 & \textbf{72.35} & 76.29 & 43.68 & 30.80 & 61.42\\
         GSM8K Metric & 81.56 & 64.06 & 58.41 & 72.23 & 76.98 & 43.73 & 32.00 & 61.28\\
         LLaMA2 Metric & 80.25 & 66.14 & \textbf{59.73} & 71.57 & \textbf{77.36} & \textbf{43.85} & 32.00 & \textbf{61.56}\\
         \bottomrule
    \end{tabular}
    }
    \label{tab:zero-2-13}
\end{table*}

\begin{table*}[htbp]
    \centering
    \caption{Accuracies (\%) of LLaMA-3 8B model for 7 zero-shot tasks with unstructured 50\% sparsity.}
    \resizebox{\textwidth}{!}{%
    \begin{tabular}{lcccccccc}
         \toprule
         Method & BoolQ & RTE & HellaSwag & WinoGrande & ARC-e & ARC-c & OBQA & Average\\
         \hline
         Dense & 81.44 & 69.68 & 60.17 & 72.85 & 80.09 & 50.43 & 34.80 & 64.21\\
         Magnitude & 49.14 & 53.43 & 38.55 & 55.09 & 60.69 & 32.42 & 24.80 & 44.87 \\
         SparseGPT & 74.80 & \textbf{54.15} & \textbf{49.90} & 68.35 & 67.05 & 36.43 & 26.40 & 53.87 \\
         Wanda & 73.43 & 52.71 & 41.80 & 63.22 & 64.86 & 29.78 & 21.80 & 49.66 \\
         RIA & 75.20 & 53.12 & 43.00 & 64.56 & 65.87 & 30.55 & 23.00 & 50.76 \\
         Pruner-Zero & 72.32 & 54.51 & 45.78 & 65.19 & 70.58 & 35.41 & 23.60 & 52.48 \\
         \hdashline
         Our Metric & \textbf{79.54} & 53.07 & 43.24 & \textbf{70.24} & \textbf{72.05} & \textbf{41.13} & \textbf{29.20} & 55.50 \\
         GSM8K Metric & 73.88 & \textbf{63.90} & 49.68 & 68.90 & 70.37 & 37.80 & 24.60 & \textbf{55.59} \\
         LLaMA3 Metric & 73.88 & \textbf{63.90} & 49.68 & 68.90 & 70.37 & 37.80 & 24.60 & \textbf{55.59} \\
         \bottomrule
    \end{tabular}
    }
    \label{tab:zero-3-8}
\end{table*}

\begin{table*}[htbp]
    \centering
    \caption{Accuracies (\%) of Instruction-tuned LLaMA-3 8B model for 7 zero-shot tasks with unstructured 50\% sparsity.}
    \resizebox{\textwidth}{!}{%
    \begin{tabular}{lcccccccc}
         \toprule
         Method & BoolQ & RTE & HellaSwag & WinoGrande & ARC-e & ARC-c & OBQA & Average\\
         \hline
         Dense & 83.06 & 67.51 & 57.68 & 71.98 & 81.61 & 52.99 & 34.20 & 64.15\\
         Magnitude & 68.84 & 60.65 & 36.31 & 53.75 & 49.83 & 26.19 & 21.80 & 45.31 \\
         SparseGPT & 77.00 & 60.65 & \textbf{49.61} & 66.46 & 70.92 & 40.19 & 26.40 & 55.89 \\
         Wanda & 76.57 & 54.51 & 41.18 & 63.61 & 67.63 & 33.70 & 22.20 & 51.34 \\
         RIA & 78.17 & 54.51 & 42.29 & 64.25 & 68.35 & 34.13 & 22.80 & 50.64 \\
         Pruner-Zero & 76.88 & 54.51 & 45.32 & 65.67 & 69.44 & 36.95 & 25.00 & 55.60 \\
         \hdashline
         Our Metric & \textbf{81.56} & 54.15  & 42.32 & \textbf{68.11} & \textbf{74.28} & \textbf{41.55} & \textbf{29.60} & \textbf{55.94}\\
         GSM8K Metric & 78.17 & 54.51 & 42.29 & 64.25 & 68.35 & 34.13 & 22.80 & 50.64 \\
         LLaMA3 Metric & 76.82 & \textbf{62.45} & 48.18 & 66.30 & 71.34 & 39.59 & 26.80 & 55.93 \\
         \bottomrule
    \end{tabular}
    }
    \label{tab:zero-3-8-it}
\vspace{2em}
    \centering
    \caption{Accuracies (\%) of Mistral 7B model for 7 zero-shot tasks with unstructured 50\% sparsity.}
    \resizebox{\textwidth}{!}{%
    \begin{tabular}{lcccccccc}
         \toprule
         Method & BoolQ & RTE & HellaSwag & WinoGrande & ARC-e & ARC-c & OBQA & Average\\
         \hline
         Dense & 81.44 & 69.68 & 60.17 & 72.85 & 80.09 & 50.43 & 34.80 & 64.21\\
         Magnitude & 75.87 & 55.60 & 56.74 & 68.35 & 74.20 & 42.15 & 27.80 & 57.24 \\
         SparseGPT & 76.73 & \textbf{61.01} & 54.52 & 67.72 & 74.24 & 41.64 & 26.60 & 57.49 \\
         Wanda & 76.12 & 55.60 & 48.95 & 65.59 & 72.69 & 37.46 & 23.00 & 54.20\\
         RIA & 76.48 & 56.68 & 49.05 & 66.30 & 72.47 & 37.12 & 22.60 & 54.39 \\
         Pruner-Zero & 77.46 & 60.65 & 50.25 & 68.90 & 71.84 & 37.46 & 22.40 & 55.57 \\
         \hdashline
         Our Metric & \textbf{82.35} & 56.68 & 55.77 & \textbf{70.88} & \textbf{76.18} & \textbf{45.22} & 28.22 & \textbf{59.33} \\
         GSM8K Metric & 81.53 & 55.60 & 54.43 & 69.38 & 74.16 & 42.15 & 26.40 & 57.66 \\
         LLaMA3 Metric & 80.52 & 56.32 & \textbf{55.94} & 69.53 & 75.00 & 42.41 & \textbf{28.40} & 58.30 \\
         \bottomrule
    \end{tabular}
    }
    \label{tab:zero-m-1}
\vspace{2em}
    \centering
    \caption{Accuracies (\%) of Instruction-tuned Mistral 7B model for 7 zero-shot tasks with unstructured 50\% sparsity.}
    \resizebox{\textwidth}{!}{%
    \begin{tabular}{lcccccccc}
         \toprule
         Method & BoolQ & RTE & HellaSwag & WinoGrande & ARC-e & ARC-c & OBQA & Average\\
         \hline
         Dense & 83.06 & 67.51 & 57.68 & 71.98 & 81.61 & 52.99 & 34.20 & 64.15\\
         Magnitude & 79.09 & 65.06 & 59.31 & 67.43 & 77.96 & 49.77 & 31.80 & 62.34 \\
         SparseGPT & 81.56 & \textbf{72.92} & 58.77 & 70.01 & 76.85 & 48.72 & 28.40 & 62.46 \\
         Wanda & 83.73 & 66.79 & 55.68 & 67.48 & 77.06 & 48.12 & 28.40 & 61.04\\
         RIA & 83.88 & 66.79 & 55.61 & 67.32 & 77.78 & 47.95 & 27.60 & 60.48 \\
         Pruner-Zero & 83.18 & 68.95 & 56.17 & 68.27 & 76.43 & 47.44 & 29.40 & 61.41 \\
         \hdashline
         Our Metric & \textbf{84.40} & 66.79 & 58.75 & \textbf{70.24} & \textbf{80.13} & \textbf{51.45} & \textbf{32.80} & \textbf{63.51} \\
         GSM8K Metric & 84.59 & 67.87 & 58.97 & 68.90 & 78.11 & 51.11 & 31.80 & 63.05 \\
         LLaMA3 Metric & 84.13 & 66.06 & \textbf{59.87} & 69.14 & 78.79 & 51.37 & 32.00 & 63.05 \\
         \bottomrule
    \end{tabular}
    }
    \label{tab:zero-m-2}
\end{table*}

\begin{table*}[htbp]
\centering
\caption{WikiText perplexity and mean zero-shot accuracies (\%) on 7 general knowledge tasks of pruned OPT models (125M to 13B) with 50\% sparsity. Best results are \textbf{bolded}.}
\label{tab:opt_part_2}
\resizebox{\textwidth}{!}{%
\begin{tabular}{lcccccccccc}
\toprule
\textbf{Model} & \textbf{Method} & \textbf{PPL} & \textbf{BoolQ} & \textbf{RTE} & \textbf{HellaSwag} & \textbf{WinoGrande} & \textbf{ARC-e} & \textbf{ARC-c} & \textbf{OBQA} & \textbf{Avg. Acc.} \\
\midrule

\multirow{4}{*}{OPT-125M} 
& SparseGPT   & 37.11 & 56.39 & 52.76 & 29.42 & 52.49 & 37.84 & 17.49 & 12.40 & 36.97 \\
& Wanda       & 38.86 & 42.45 & 52.71 & 27.98 & 51.62 & 35.40 & 16.64 & 12.00 & 34.11 \\
& RIA         & 38.97 & 40.61 & 52.35 & 27.63 & 52.25 & 34.85 & 17.06 & 12.20 & 33.85 \\
& \textsc{OptiShear}       & 36.81 & 58.78 & 52.71 & 29.11 & 52.96 & 36.62 & 17.41 & 12.40 & \textbf{37.14} \\
\midrule

\multirow{4}{*}{OPT-350M} 
& SparseGPT   & 34.72 & 57.28 & 50.54 & 28.01 & 52.72 & 38.30 & 19.54 & 13.60 & 37.14 \\
& Wanda       & 35.97 & 60.09 & 53.43 & 27.69 & 52.09 & 36.28 & 17.66 & 11.00 & 36.89 \\
& RIA         & 35.82 & 58.62 & 54.51 & 27.60 & 51.85 & 35.65 & 18.17 & 11.20 & 36.80 \\
& \textsc{OptiShear}       & 34.21 & 62.02 & 52.35 & 27.99 & 50.75 & 37.46 & 19.37 & 14.20 & \textbf{37.73} \\
\midrule

\multirow{4}{*}{OPT-1.3B} 
& SparseGPT   & 34.54 & 56.21 & 50.79 & 35.86 & 58.80 & 50.80 & 21.84 & 18.00 & 41.76 \\
& Wanda       & 35.93 & 59.63 & 54.15 & 30.62 & 54.14 & 41.84 & 17.41 & 12.80 & 38.66 \\
& RIA         & 35.81 & 60.55 & 56.68 & 30.64 & 52.64 & 42.13 & 17.49 & 12.60 & 38.96 \\
& \textsc{OptiShear}       & 34.72 & 55.50 & 50.18 & 36.32 & 57.30 & 51.09 & 22.78 & 18.00 & \textbf{41.60} \\
\midrule

\multirow{4}{*}{OPT-2.7B} 
& SparseGPT   & 13.51 & 62.23 & 52.35 & 40.69 & 56.75 & 54.59 & 24.49 & 18.40 & 44.27 \\
& Wanda       & 14.39 & 62.26 & 52.71 & 32.08 & 50.99 & 44.19 & 18.69 & 14.40 & 39.33 \\
& RIA         & 14.02 & 62.29 & 52.71 & 31.78 & 50.83 & 44.11 & 19.11 & 14.80 & 39.38 \\
& \textsc{OptiShear}       & 13.17 & 63.79 & 51.26 & 40.45 & 56.27 & 55.30 & 24.83 & 18.80 & \textbf{44.39} \\
\midrule

\multirow{4}{*}{OPT-6.7B} 
& SparseGPT   & 11.51 & 65.90 & 51.99 & 44.14 & 60.96 & 60.61 & 26.71 & 22.40 & 47.53 \\
& Wanda       & 12.05 & 62.14 & 52.71 & 33.62 & 52.49 & 50.13 & 19.97 & 14.60 & 40.81 \\
& RIA         & 11.74 & 62.26 & 52.71 & 33.80 & 54.78 & 49.92 & 20.31 & 14.20 & 41.14 \\
& \textsc{OptiShear}       & 11.32 & 65.08 & 54.15 & 44.06 & 60.93 & 60.61 & 26.11 & 22.20 & \textbf{47.59} \\
\midrule

\multirow{4}{*}{OPT-13B} 
& SparseGPT   & 11.28 & 61.74 & 57.40 & 45.06 & 63.06 & 62.54 & 29.18 & 21.80 & 48.68 \\
& Wanda       & 11.56 & 65.63 & 53.07 & 37.17 & 56.99 & 52.10 & 22.61 & 16.40 & 43.42 \\
& RIA         & 11.43 & 64.95 & 52.71 & 36.80 & 57.85 & 53.03 & 22.53 & 16.40 & 43.47 \\
& \textsc{OptiShear}       & 11.15 & 62.45 & 59.57 & 48.17 & 64.33 & 61.83 & 29.61 & 22.80 & \textbf{49.82} \\
\bottomrule
\end{tabular}
}
\end{table*}

\paragraph{Speedup Analysis}
Our meta pruning metric has a theoretical complexity of $O(d_{hidden}^2)$, compared to $O(d_{hidden}^3)$ for SparseGPT. We measure empirical pruning time on NVIDIA RTX A6000 GPUs at 50\% sparsity, using C4 calibration data for activation estimation.
As shown in Table \ref{tab:prune-speed}, our meta pruning metric results in negligible time overhead compared to SparseGPT. We further evaluate the inference speedup for semi-structured 4:8 and 2:4 sparsity on NVIDIA RTX A6000 GPUs.
Our simulations utilize the high-performance GEMM kernel from the NVIDIA CUTLASS library. According to the results presented in Table \ref{tab:infer-speed}, when compared with dense models, we observe an average speedup of 1.20$\times$ in end-to-end latency.

\begin{table}[htbp]
    \begin{minipage}[h]{0.5\textwidth}
    \fontsize{9}{8} \selectfont
    \caption{Pruning speed for pruning LLaMA-2/3 and Mistral models to 50\% sparsity.}
    \begin{tabular}{lcccc}
    \toprule
        Method & L2-7B & L2-13B & L3-8B & M-7B \\
     \hline
        SparseGPT  & 370.03 & 464.77 & 457.71 &  450.76   \\
        \rowcolor{gray!30}
        \textsc{\textsc{OptiShear}} & \textbf{56.16} & \textbf{107.11} & \textbf{60.11} &\textbf{59.80}  \\
    \bottomrule
    \end{tabular}
    \label{tab:prune-speed}
    \end{minipage}
    \hspace{3mm}
    \begin{minipage}[h]{0.48\textwidth}
        \fontsize{9}{8} \selectfont
        \caption{Inference speedup of different sparsity patterns for LLaMA-2/3 and Mistral models.}
        \begin{tabular}{lcccc}
        \toprule
        Sparsity & L2-7B & L2-13B & L3-8B & M-7B \\
     \hline
        4:8 & {1.11}$\times$ & 1.04$\times$ & 1.15$\times$ & \textbf{1.17}$\times$ \\
        2:4 & \textbf{1.35}$\times$ & \textbf{1.14}$\times$ & \textbf{1.15}$\times$ & 1.16$\times$ \\
    \bottomrule
    \end{tabular}
    \label{tab:infer-speed}
    \end{minipage}
\end{table}

\begin{figure*}[htbp]
    \centering
    \makebox[\textwidth][c]{
        \begin{minipage}{1.0\textwidth} 
            \centering
            \begin{subfigure}[b]{0.3\textwidth}
                \centering
                \includegraphics[width=\textwidth]{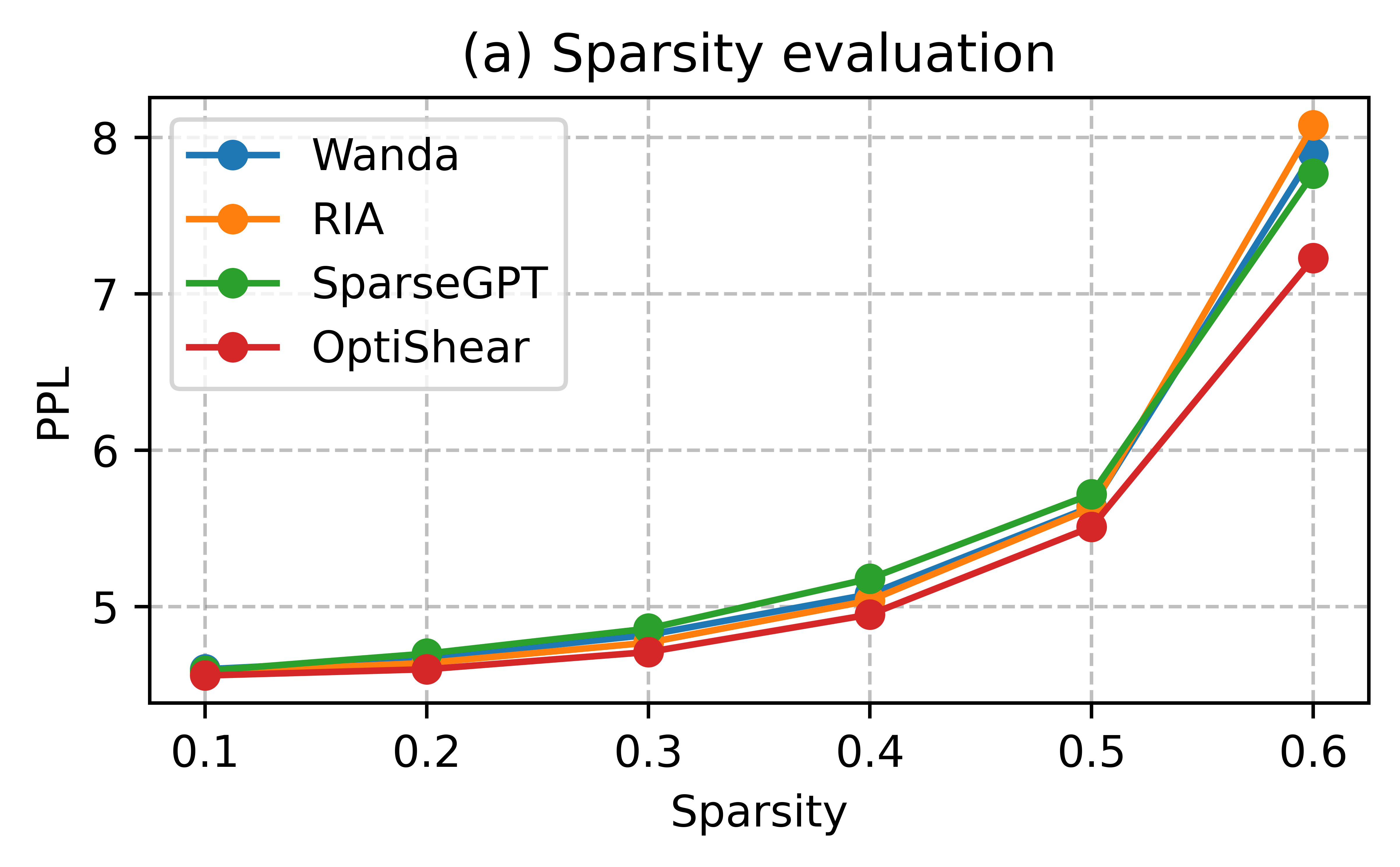}
                \label{fig:sparsity}
            \end{subfigure}
            \hfill
            \begin{subfigure}[b]{0.3\textwidth}
                \centering
                \includegraphics[width=\textwidth]{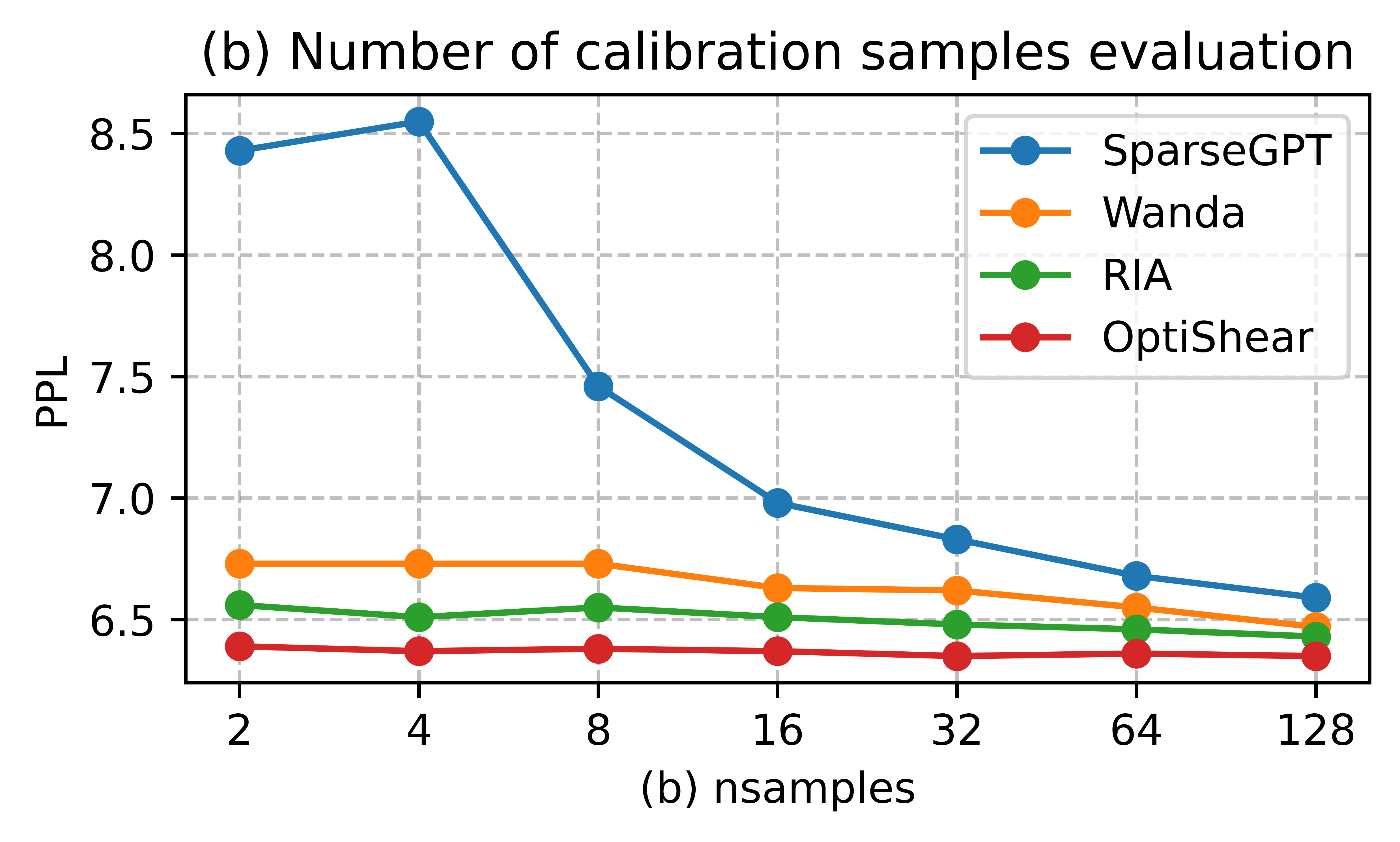}
                \label{fig:nsample}
            \end{subfigure}
            \hfill
            \begin{subfigure}[b]{0.3\textwidth}
                \centering
                \includegraphics[width=\textwidth]{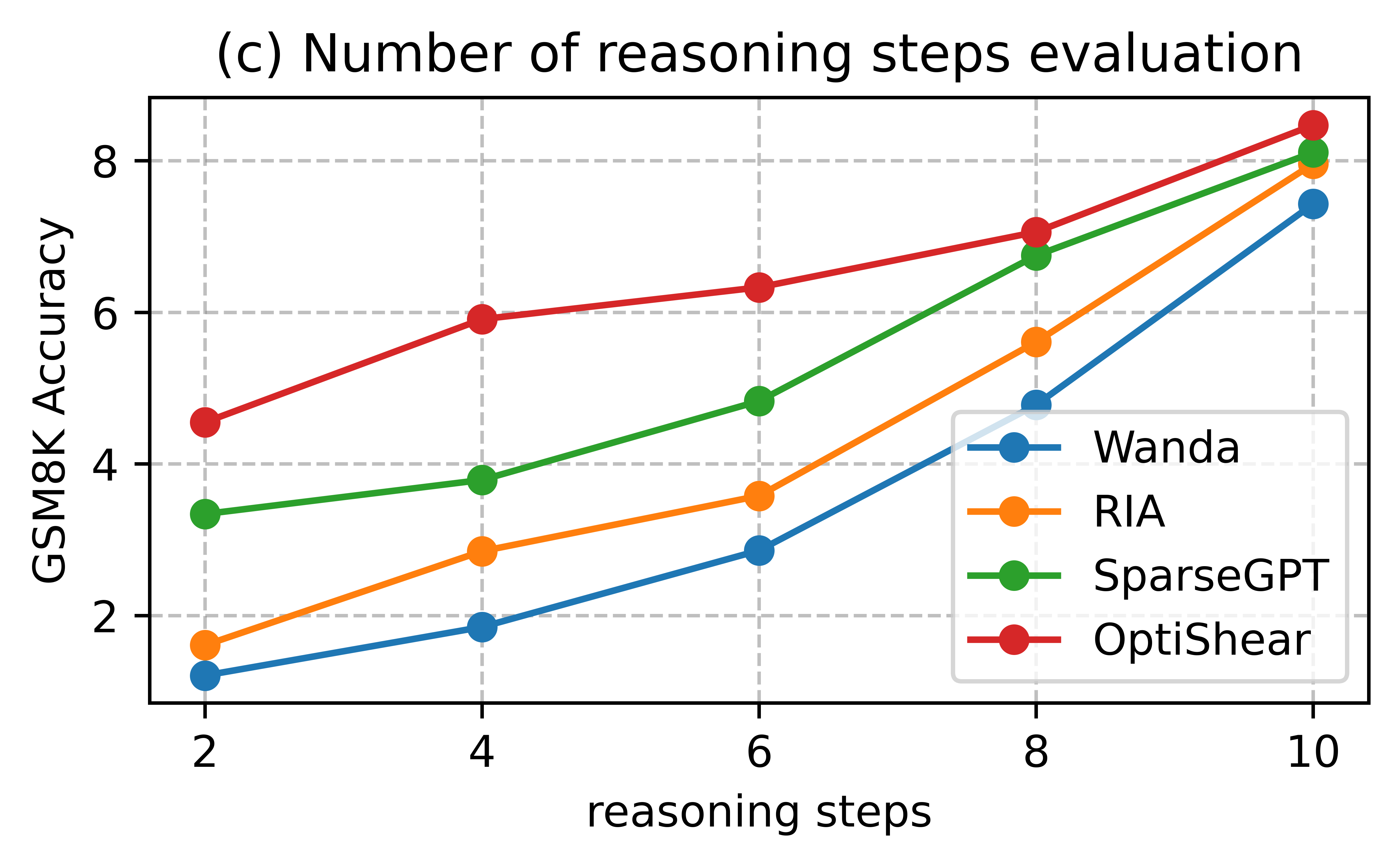}
                \label{fig:cot_steps}
            \end{subfigure}
            \vspace{-1em}
            \caption{Sensitivity evaluation across sparsity levels, calibration data size, and reasoning steps in calibration samples for mathematical reasoning.}
            \label{fig:sensi_eval}
        \end{minipage}
    }
\end{figure*}

\vspace{-1em}
\paragraph{Different Sparsity \& Calibration Samples.} 
In Figure~\ref{fig:sensi_eval}(a), we evaluate LLaMA-2 13B under varying sparsity ratios (0.1–0.6). \textsc{OptiShear} consistently outperforms all baselines, achieving a 10.52\% relative improvement over RIA at 60\% sparsity.
Figure~\ref{fig:sensi_eval}(b) examines the impact of calibration sample size (ranging from 2 to 128) on LLaMA-2 7B. While SparseGPT's performance fluctuates with sample size, \textsc{OptiShear} remains robust and consistently superior, even with very limited calibration data.
In the case of mathematical reasoning (Figure~\ref{fig:sensi_eval}(c)), increasing the number of reasoning steps in calibration samples improves performance across all methods. \textsc{OptiShear}, however, shows consistently better results, demonstrating its effectiveness under diverse reasoning conditions.

\paragraph{Optimal pruning metrics.}
In Table\ref{tab:zero-metrics}, we present the optimal coefficients and transformation operations for pruning metrics obtained using calibration data sampled from the C4 dataset. In contrast, Table \ref{tab:gsm8k-metric} displays the corresponding optimal coefficients and operations when using samples from the GSM8K dataset as calibration data. When compared to the results based on the C4 dataset, the pruning metrics derived from the GSM8K dataset exhibit a greater degree of divergence from the commonly used RIA metric \citep{zhangplug}. Notably, the majority of these optimized metrics do not utilize the relative sum as a weight coefficient, suggesting a distinct preference in metric structure when calibrated on complex reasoning tasks.

\begin{table}[htbp]
\caption{Optimal coefficients and operations for pruning metrics on C4 calibration data. }
\resizebox{\textwidth}{!}{%
\begin{tabular}{lcccccccc}
\toprule
\multirow{2}{*}{Metric} &
  \multicolumn{2}{c}{LLaMA-1} &
  \multicolumn{2}{c}{LLaMA-2} &
  \multicolumn{2}{c}{LLaMA-3} &
  \multicolumn{2}{c}{Mistral} \\      & 7B    & 13B    & 7B    & 13B   
          & 8B    & 8B-Inst
           & 7B    & 7B-Inst \\
           \hline
$\alpha$       & relative sum & relative sum & relative sum & relative sum & relative sum & no coe & relative sum & F norm   \\
$\beta$   & to mean &  to mean & to mean & no coe & F norm & F norm & to mean & to mean   \\
$\tau_{1}$  & no op &  square & no op & square & no op & no op & square & sqrt   \\
$\tau_{2}$  & sqrt & no op  & sqrt & sqrt & no op & no op & no op & sqrt   \\
\bottomrule
\end{tabular}%
}
\label{tab:zero-metrics}
\end{table}

\begin{table}[htbp]
\caption{Optimal coefficients and operations for pruning metrics on GSM8K calibration data. }
\resizebox{\textwidth}{!}{%
\begin{tabular}{lcccccccc}
\toprule
\multirow{2}{*}{Metric} &
  \multicolumn{2}{c}{LLaMA-1} &
  \multicolumn{2}{c}{LLaMA-2} &
  \multicolumn{2}{c}{LLaMA-3} &
  \multicolumn{2}{c}{Mistral} \\      & 7B    & 13B    & 7B    & 13B   
          & 8B    & 8B-Inst
           & 7B    & 7B-Inst \\
           \hline
$\alpha$       & row sum & to mean & F norm & column sum & to mean & relative sum & row sum & relative sum  \\
$\beta$   & relative sum &  F norm & to sum & relative sum & to sum & no coe & no coe & to mean  \\
$\tau_{1}$  & no op & no op  & no op & square & square & no op & square & square   \\
$\tau_{2}$  & sqrt & sqrt  & sqrt & sqrt & sqrt & no op & no op & no op   \\
\bottomrule
\end{tabular}%
}
\label{tab:gsm8k-metric}
\end{table}

\subsection{Appendix B: Method Details}
\label{app:method_details}

\paragraph{\textsc{OptiShear} Algorothim}

\textsc{OptiShear} employs an evolutionary search framework to automatically discover adaptive pruning metrics tailored to specific model-task pairs. As outlined in Algorithm~\ref{alg:search}, we define a flexible metric space combining weight magnitudes and input activations, and evolve configurations using NSGA-II~\cite{deb2002fast}. Each candidate metric is evaluated by applying it across all layers, retaining the top-$k\%$ weights per layer according to the target sparsity. The resulting pruned model is scored using a global divergence objective $\mathcal{L}_{\text{div}}$, which measures output discrepancy on a calibration dataset. Through iterative selection, crossover, and mutation, our method converges to high-performing metric configurations that balance weight-activation interaction for effective pruning.

\begin{algorithm}[htbp]
\small
\caption{\textsc{OptiShear}: Evolutionary Search for Adaptive Pruning Metrics}
\label{alg:search}
\begin{algorithmic}[1]
    \State \textbf{Input:} Model $f$, Calibration dataset $\mathcal{D}$, Target sparsity $s_{\text{target}}$
    \State \textbf{Output:} Best metric configuration $(\alpha^*, \beta^*, F_1^*, F_2^*)$ 
    
    \State Initialize population with random $(\alpha, \beta, F_1, F_2)$ 
        \hfill $\rhd$~\textcolor{blue}{\textit{Metric search space initialization}}
    
    \Repeat
        \For{each individual}
            \State Apply metric and prune top-$k\%$ weights per layer 
            \State Compute $\mathcal{L}_{\text{div}}$ on $\mathcal{D}$ as fitness
                \hfill $\rhd$~\textcolor{blue}{\textit{Measures global consistency after pruning}}
        \EndFor
        
        \State Select parents via crowding distance
            \hfill $\rhd$~\textcolor{blue}{\textit{Population-based exploration}}
        \State Generate offspring using SBX and polynomial mutation

    \Until{converged or max generations reached}
        \hfill $\rhd$~\textcolor{blue}{\textit{Early stopping if converged}}
        
    \State \textbf{return} Best configuration
\end{algorithmic}
\end{algorithm}

\paragraph{Coefficient and Transformation Function Definitions.}
To provide a precise understanding of the components in our meta pruning metric, we present the mathematical formulations of the coefficient functions ($\alpha(\cdot), \beta(\cdot)$) and transformation functions ($F_1(\cdot), F_2(\cdot)$) in Table \ref{tab:coe_eqs} and Table \ref{tab:transformation_candidates}. The coefficient functions are used to normalize raw weight magnitudes or activation norms, while the transformation functions introduce nonlinearity and adaptive reweighting. These definitions correspond to the candidate operations listed in Table~\ref{coe-ope}, and together they define the search space explored by \textsc{OptiShear}.

\begin{table*}[t!]
\centering
\caption{Coefficient functions used in the meta pruning metric.}
\label{tab:coe_eqs}
\resizebox{0.95\textwidth}{!}{%
\begin{tabular}{lcc}
\toprule
\textbf{Name} & \textbf{Weight Coefficient ($\alpha(|W_{ij}|)$)} & \textbf{Activation Coefficient ($\beta(\|X_j\|_2)$)} \\
\midrule
Uniform Weighting         & $ 1 $                            & $ 1 $ \\
Global Sum Scaling        & $ \left( \sum_{i,j} |W_{ij}| \right)^{-1} $ & $ \left( \sum_{j} \|X_j\|_1 \right)^{-1} $ \\
Frobenius Normalization   & $ \|W\|_F^{-1} $                & $ \|X\|_F^{-1} $ \\
Global Mean Scaling       & $ mn \left( \sum_{i,j} |W_{ij}| \right)^{-1} $ & $ n \left( \sum_{j} \|X_j\|_1 \right)^{-1} $ \\
Row-wise Normalization    & $ \left( \sum_j |W_{ij}| \right)^{-1} $ for each row $i$ & $ \left( \sum_t |X_{tj}| \right)^{-1} $ for each neuron $j$ \\
Column-wise Normalization & $ \left( \sum_i |W_{ij}| \right)^{-1} $ for each column $j$ & $ \left( \sum_t |X_{tj}| \right)^{-1} $ for each token position $t$ \\
Relative Magnitude Scaling & $ \text{row-sum}(|W_{ij}|) + \text{column-sum}(|W_{ij}|) $ & $ \text{row-sum}(\|X_j\|_2) + \text{column-sum}(\|X_j\|_2) $ \\
\bottomrule
\end{tabular}
}
\end{table*}

\begin{table*}[htbp]
\centering
\caption{Candidate transformation functions applied to weights and activations in the meta pruning metric.}
\label{tab:transformation_candidates}
\resizebox{0.9\textwidth}{!}{%
\begin{tabular}{lcc}
\toprule
\textbf{Name} & \textbf{Weight Transformation ($F_1(|W_{ij}|)$)} & \textbf{Activation Transformation ($F_2(\|X_j\|_2)$)} \\
\midrule
Identity        & $ |W_{ij}| $                          & $ \|X_j\|_2 $ \\
Square          & $ |W_{ij}|^2 $                        & $ \|X_j\|_2^2 $ \\
Square Root     & $ \sqrt{|W_{ij}|} $                   & $ \sqrt{\|X_j\|_2} $ \\
Logarithm       & $ \log(1 + |W_{ij}|) $                & $ \log(1 + \|X_j\|_2) $ \\
Exponential     & $ \exp(-|W_{ij}|) $                    & $ \exp(-\|X_j\|_2) $ \\
Sigmoid         & $ \sigma(|W_{ij}|) $                 & $ \sigma(\|X_j\|_2) $ \\
Softmax         & $ \mathrm{Softmax}_i(|W_{ij}|) $      & $ \mathrm{Softmax}_j(\|X_j\|_2) $ \\
\bottomrule
\end{tabular}
}
\end{table*}

\begin{figure*}[htbp]
    \centering
    \begin{minipage}{0.33\linewidth}
        \centering
        \includegraphics[width=\textwidth]{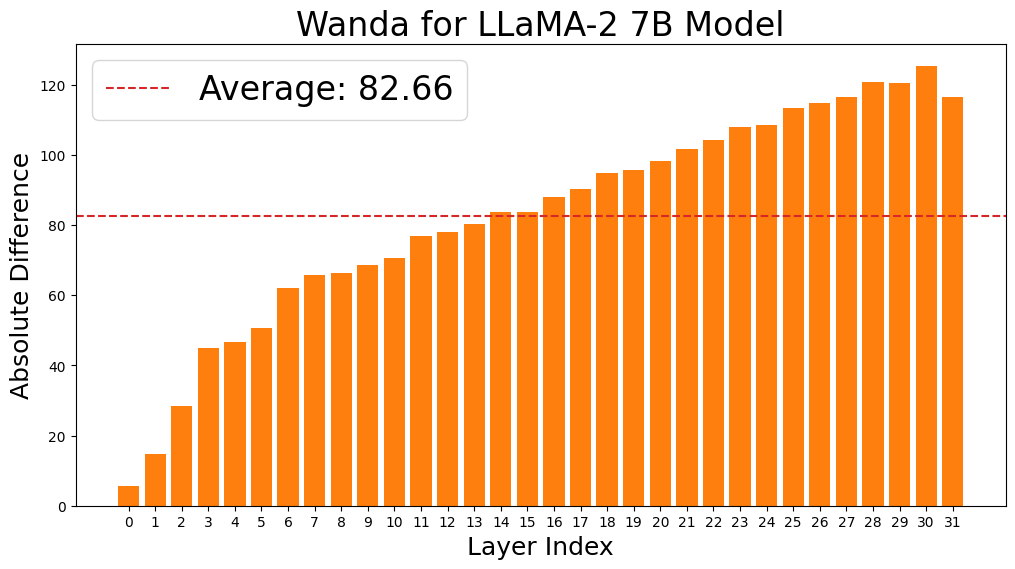}
    \end{minipage}\hfill
    \begin{minipage}{0.33\linewidth}
        \centering
        \includegraphics[width=\textwidth]{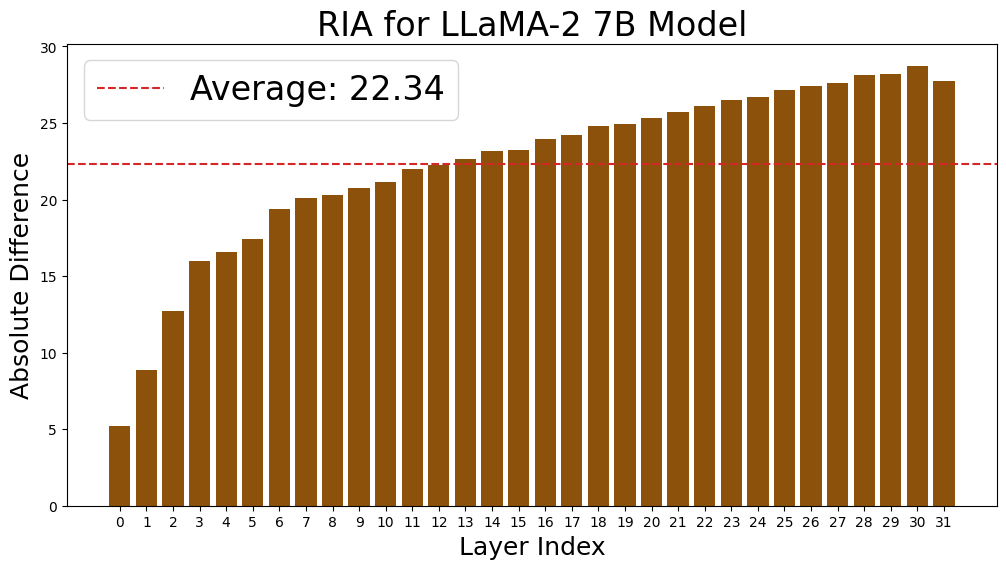} 
    \end{minipage}\hfill
    \begin{minipage}{0.33\linewidth}
        \centering
        \includegraphics[width=\textwidth]{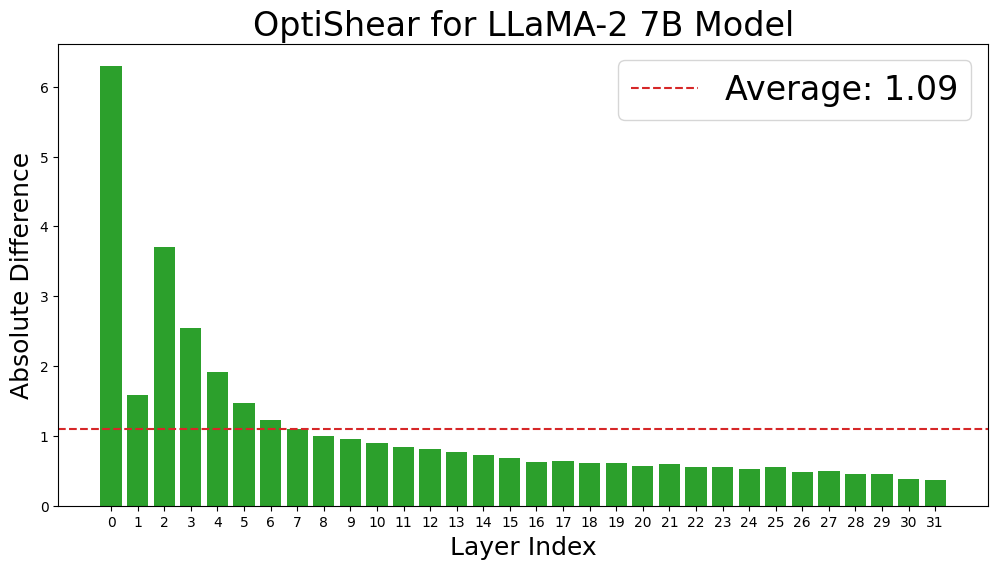} 
    \end{minipage}
    \caption{Layerwise absolute distance between transformed weights and transformed activations for Wanda, RIA, and OptiShear metrics on LLaMA-2 7B models.}
    \label{fig:llama2_7b_distance}
\end{figure*}

\begin{figure*}[htbp]
    \centering
    \begin{minipage}{0.33\linewidth}
        \centering
        \includegraphics[width=\textwidth]{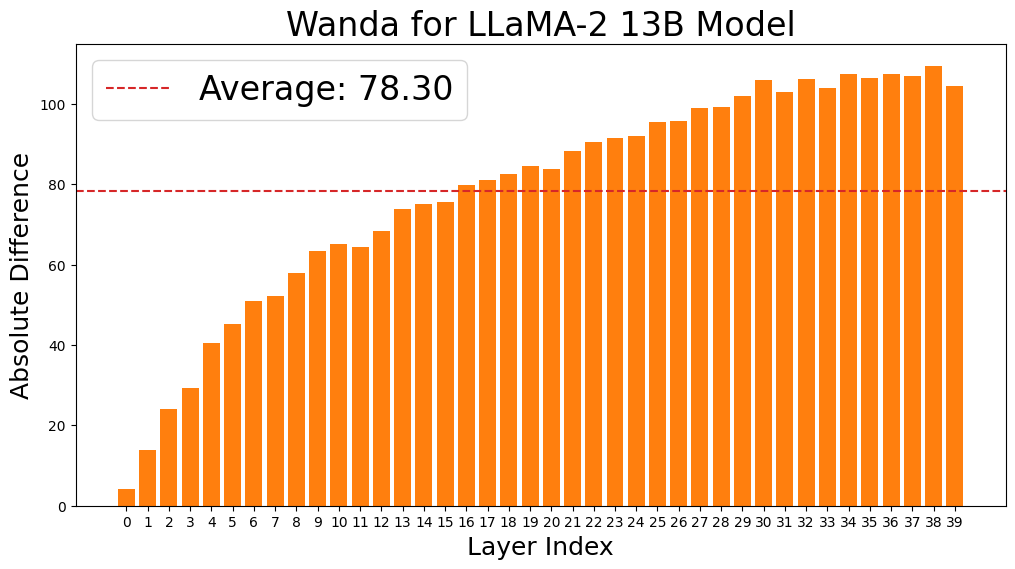}
    \end{minipage}\hfill
    \begin{minipage}{0.33\linewidth}
        \centering
        \includegraphics[width=\textwidth]{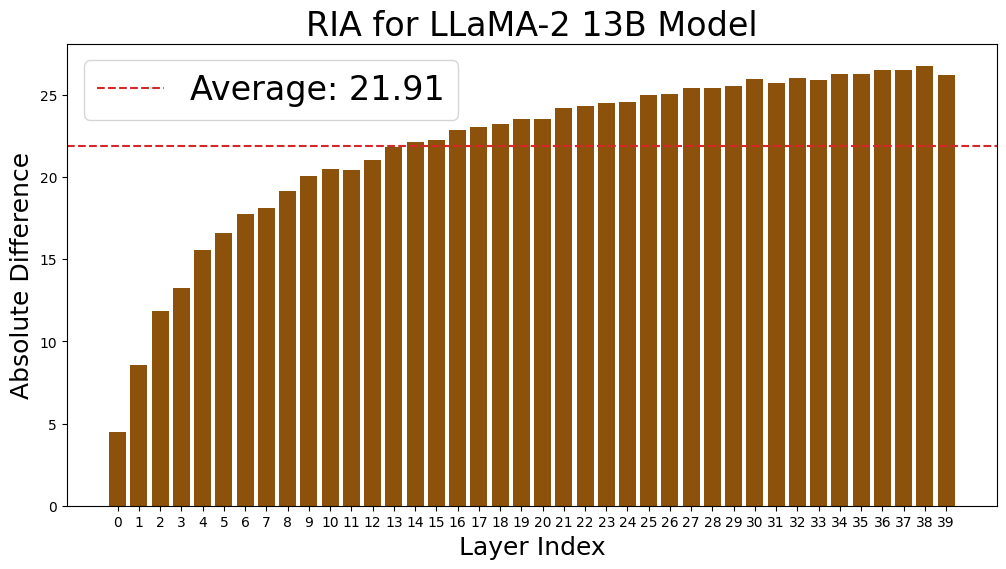} 
    \end{minipage}\hfill
    \begin{minipage}{0.33\linewidth}
        \centering
        \includegraphics[width=\textwidth]{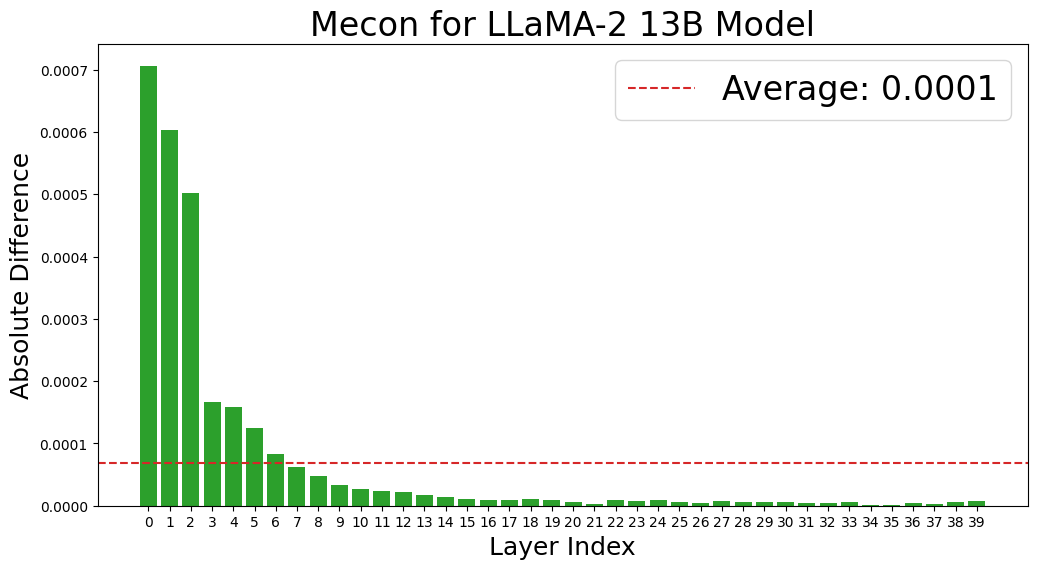} 
    \end{minipage}
    \caption{Layerwise absolute distance between transformed weights and transformed activations for Wanda, RIA, and OptiShear metrics on LLaMA-2 13B models.}
    \label{fig:llama2_13b_distance}
\end{figure*}

\begin{figure*}[htbp]
    \centering
    \begin{minipage}{0.33\linewidth}
        \centering
        \includegraphics[width=\textwidth]{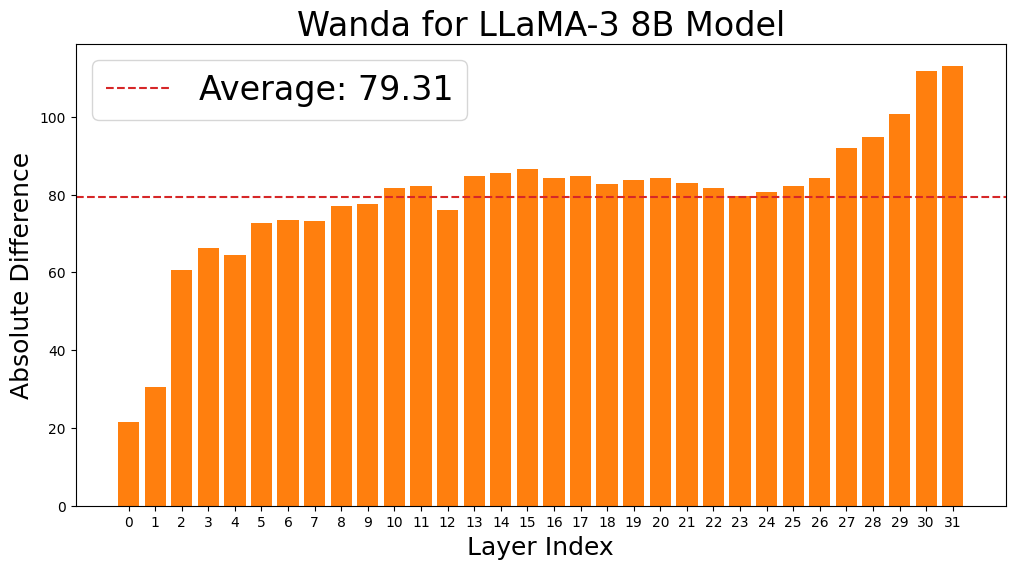}
    \end{minipage}\hfill
    \begin{minipage}{0.33\linewidth}
        \centering
        \includegraphics[width=\textwidth]{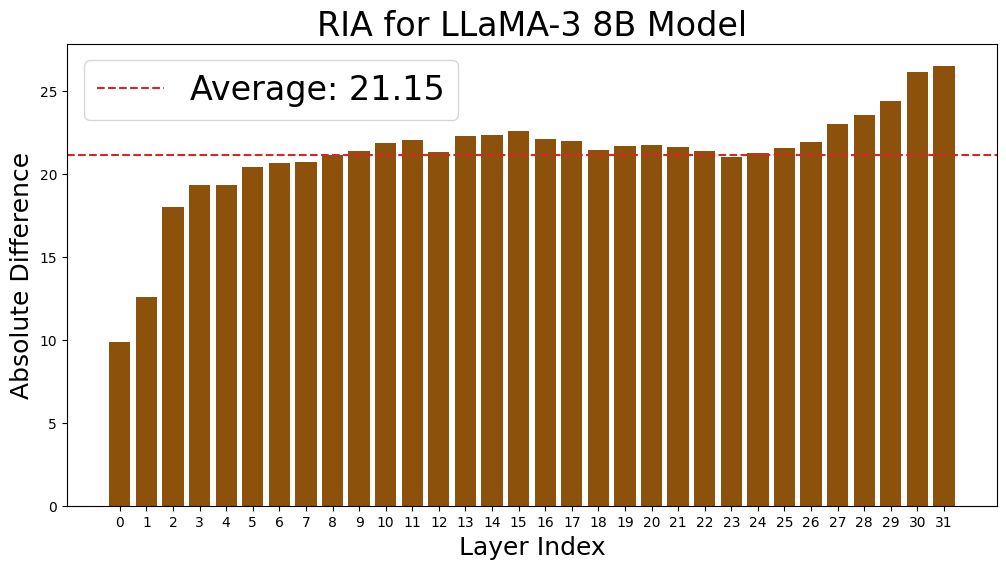} 
    \end{minipage}\hfill
    \begin{minipage}{0.33\linewidth}
        \centering
        \includegraphics[width=\textwidth]{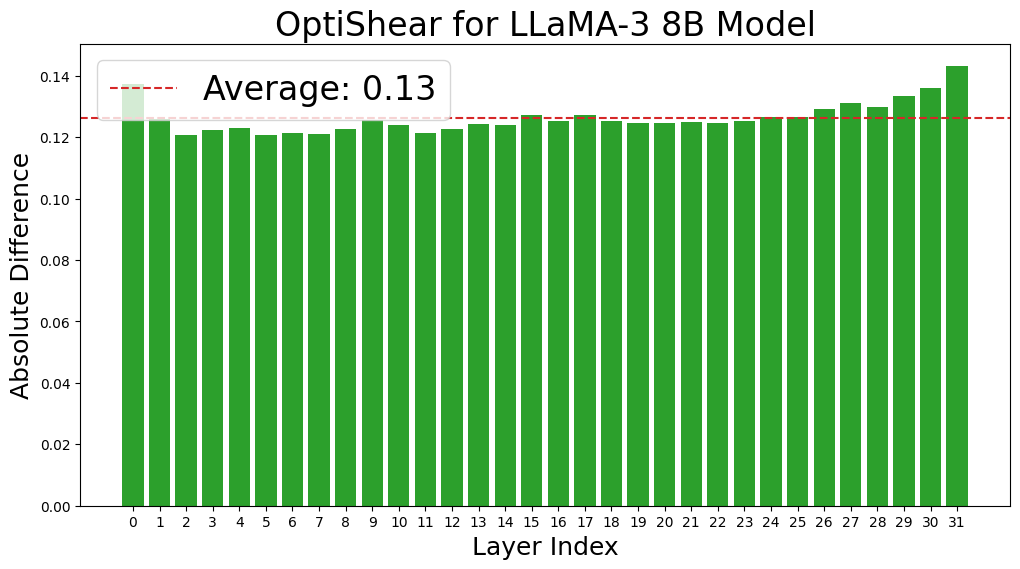} 
    \end{minipage}
    \caption{Layerwise absolute distance between transformed weights and transformed activations for Wanda, RIA, and OptiShear metrics on LLaMA-3 8B models.}
    \label{fig:llama3_8b_distance}
\end{figure*}

\begin{figure*}[htbp]
    \centering
    \begin{minipage}{0.33\linewidth}
        \centering
        \includegraphics[width=\textwidth]{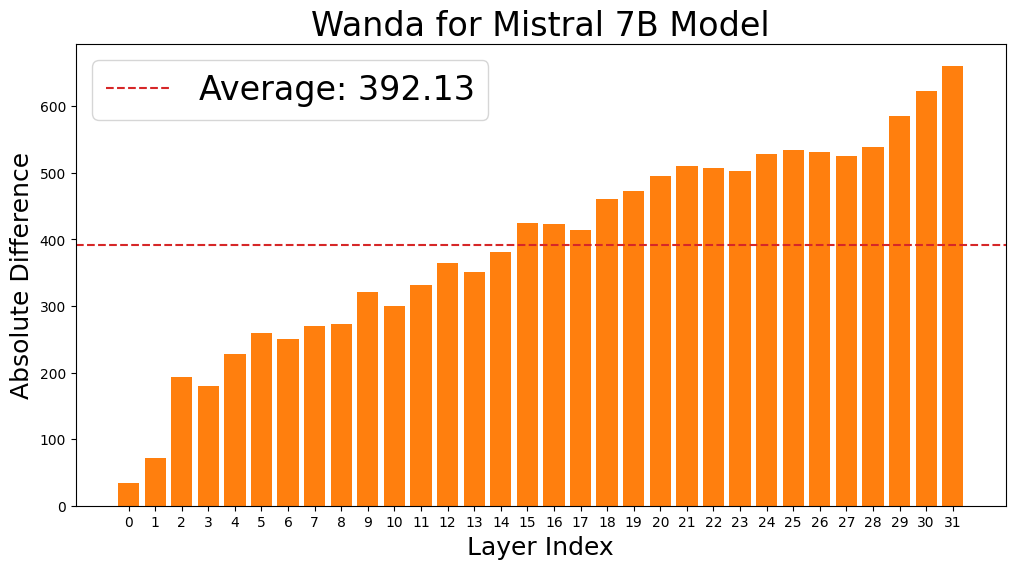}
    \end{minipage}\hfill
    \begin{minipage}{0.33\linewidth}
        \centering
        \includegraphics[width=\textwidth]{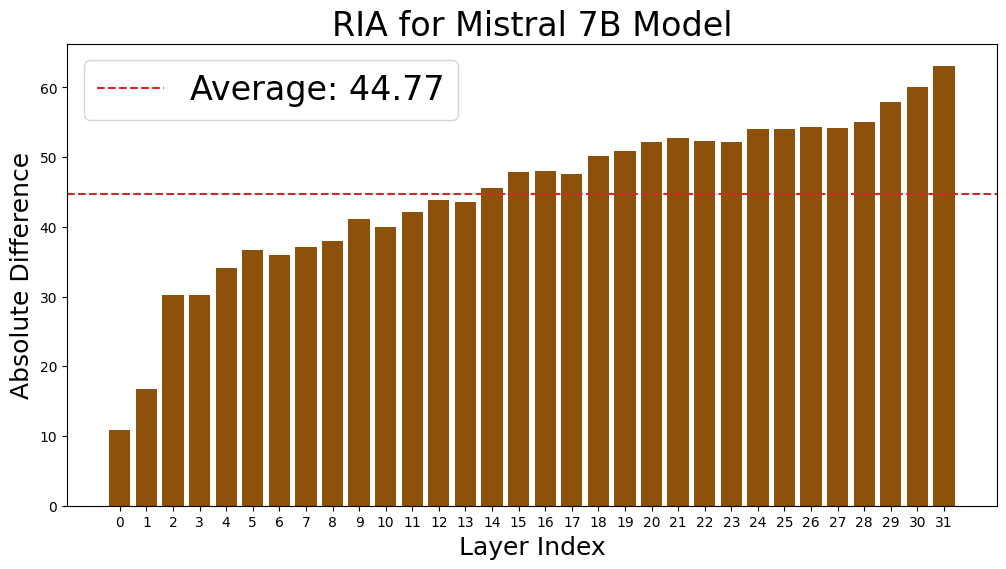} 
    \end{minipage}\hfill
    \begin{minipage}{0.33\linewidth}
        \centering
        \includegraphics[width=\textwidth]{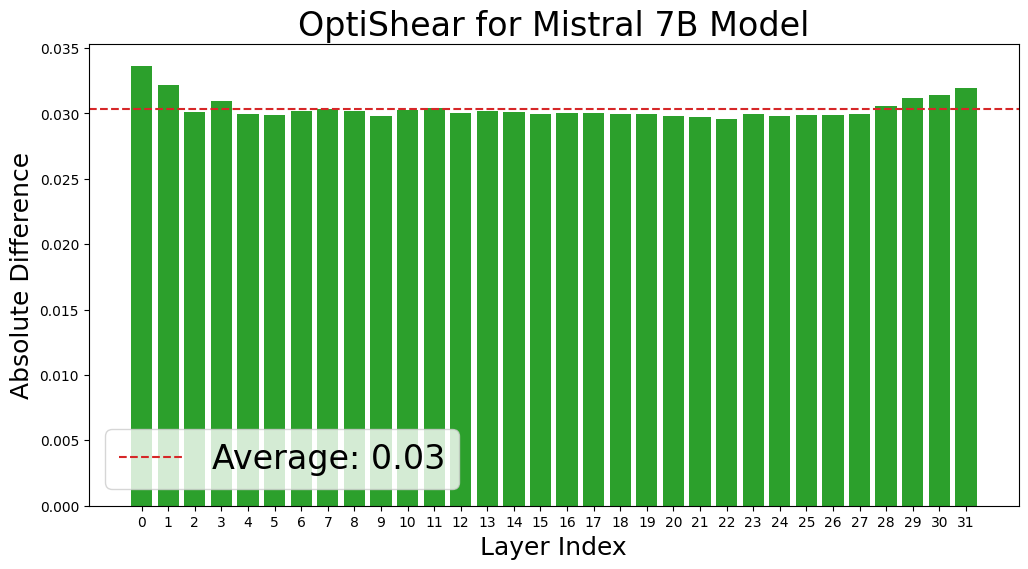} 
    \end{minipage}
    \caption{Layerwise absolute distance between transformed weights and transformed activations for Wanda, RIA, and OptiShear metrics on Mistral 7B models.}
    \label{fig:mistral_7b_distance}
\end{figure*}

\paragraph{Layer-wise weight-activation alignment.}
\label{sec:w_x}
In our analysis of the optimal searched pruning metrics, we provide with detailed layer-wise difference curves available in Figures \ref{fig:llama2_7b_distance}, \ref{fig:llama2_13b_distance}, \ref{fig:llama3_8b_distance}, 
\ref{fig:mistral_7b_distance}.

\paragraph{Search Trial Number Analysis.}

\begin{figure}[htbp]
    \centering
    \includegraphics[width=0.5\linewidth]{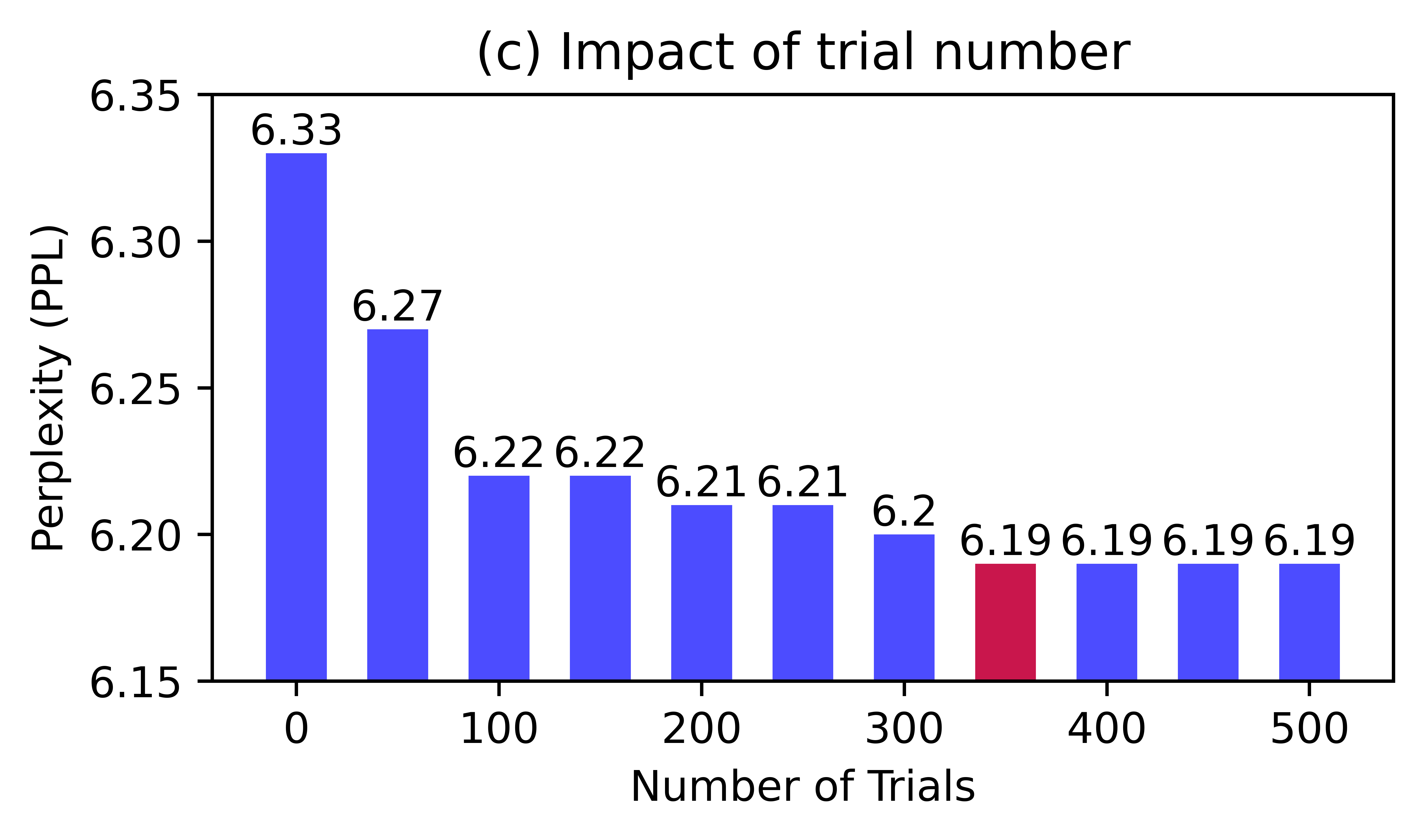}
    \caption{Performance trend as the number of search trials increases.}
    \label{fig:ntrials}
\end{figure}

Using LLaMA-2 7B pruned to 50\% unstructured sparsity on the WikiText perplexity task, we evaluate the impact of the number of search trials. We run a total of 500 trials and observe that performance converges around trial 350.

\subsection{Limitations} \label{sec:limitation}
While \textsc{OptiShear} demonstrates promising results in adaptive pruning for large language models, there are several limitations to consider.
(1) Although our method identifies the optimal pruning metrics through a search process and experimentally validates the potential factors affecting pruning metrics across different distribution models, a universal pruning metric that is both theoretically sound and empirically effective for all distribution models remains unexplored.
(2) Furthermore, \textsc{OptiShear} shows reduced effectiveness on LLaMA-3 models under semi-structured pruning constraints (4:8 and 2:4 sparsity patterns), likely due to the model's higher knowledge density making it more sensitive to parameter block removal. Developing more sophisticated structured pruning techniques for such knowledge-dense models represents a promising direction for future research.
\end{document}